\documentclass[sigconf]{acmart}

\renewcommand\footnotetextcopyrightpermission[1]{} 
\AtBeginDocument{%
  \providecommand\BibTeX{{%
    \normalfont B\kern-0.5em{\scshape i\kern-0.25em b}\kern-0.8em\TeX}}}

\acmConference[ACM FAccT '21]{ACM Conference on Fairness, Accountability, and Transparency}{March 03--10, 2021}{}

\usepackage{shortcuts}
\usepackage{amsmath, latexsym}
\usepackage{amscd}
\usepackage{amsbsy}
\usepackage{amsthm}
\usepackage{comment} 
\usepackage{nicefrac}

\usepackage{enumitem}

\newtheorem{theorem}{Theorem}
\newtheorem{lemma}{Lemma}
\newtheorem{corollary}{Corollary}
\theoremstyle{definition}
\newtheorem{assumption}{Assumption}
\newtheorem{definition}{Definition}

\newtheorem{proposition}{Proposition}
\usepackage{url}
\usepackage{outlines}
\newcommand{\pr}{\mathbb{P}}
\usepackage{empheq}
\usepackage{cleveref} 
\usepackage{xcolor}

\usepackage{pifont}%

\makeatletter
\newcommand*{\defeq}{\mathrel{\rlap{%
			\raisebox{0.3ex}{$\m@th\cdot$}}%
		\raisebox{-0.3ex}{$\m@th\cdot$}}%
	=}
\makeatother

\newcommand{\propa}{\rho_a}
\newcommand{\propb}{\rho_b}
\newcommand{\cmark}{\ding{51}}%
\newcommand{\xmark}{\ding{55}}%

\newcommand{\Aa}{\mathcal A} 

\newcommand{\heta}{ \hat D}
\newcommand{\hp}{ \hat p^*}
\newcommand{\Dbarx}{\overline{D}(x)}

\newcommand{\Dbarxa}{\overline{D}(x,a)}

\newcommand{\lxa}{\lambda^*_{xa}}
\newcommand{\lx}{\lambda^*_{x}}
\newcommand{\xiax}{\E[\xi(A)\mid X=x]}
\newcommand{\bexpx}{\E[\beta_A\mid X=x] }
\newcommand{\pxa}{p^*(x,a)}
\newcommand{\px}{p^*(x)}
\newcommand{\Rev}{{R}}
\newcommand{\pstar}{p^*}

\newcommand{\Dmnd}{D}

\usepackage[export]{adjustbox}%

\usepackage{outlines}
\usepackage{tikz} 
\usetikzlibrary{shapes,decorations,arrows,calc,arrows.meta,fit,positioning}
\tikzset{
    -Latex,auto,node distance =1 cm and 1 cm,semithick,
    state/.style ={circle, draw, minimum width = 0.7 cm},
    point/.style = {circle, draw, inner sep=0.04cm,fill,node contents={}},
    bidirected/.style={Latex-Latex,dashed},
    el/.style = {inner sep=2pt, align=left, sloped}
}
\usepackage{caption}
\usepackage{subcaption}

\usepackage{array}
\usepackage{makecell}

\title{Fairness, Welfare, and Equity in Personalized Pricing}
\author{Nathan Kallus}
\affiliation{Cornell University and Cornell Tech}
\email{ kallus@cornell.edu}
\author{Angela Zhou*}
\affiliation{Cornell University and Cornell Tech}
 \email{az434@cornell.edu}

\begin{abstract}
    We study the interplay of fairness, welfare, and equity considerations in personalized pricing based on customer features.
Sellers are increasingly able to conduct \textit{price personalization} based on predictive modeling of demand conditional on covariates: setting customized interest rates, targeted discounts of consumer goods, and personalized subsidies of scarce resources with positive externalities like vaccines and bed nets. These different application areas may lead to  \textit{different} concerns around fairness, welfare, and equity on \emph{different} objectives: price burdens on consumers, price envy, firm revenue, access to a good, equal access, and distributional consequences when the good in question further impacts downstream outcomes of interest. We conduct a comprehensive literature review in order to disentangle these different normative considerations and propose a taxonomy of different objectives with mathematical definitions.
    We focus on observational metrics that do not assume access to an underlying valuation distribution which is either unobserved due to binary feedback or ill-defined due to overriding behavioral concerns regarding interpreting revealed preferences. 
    In the setting of personalized pricing for the provision of goods with positive benefits, we discuss how price optimization may provide unambiguous benefit by achieving a ``triple bottom line'': personalized pricing enables \textit{expanding access}, which in turn may lead to \textit{gains in welfare} due to heterogeneous utility, and improve \textit{revenue or budget utilization}.  We empirically demonstrate the potential benefits of personalized pricing in two settings: pricing subsidies for an elective vaccine, and the effects of personalized interest rates on downstream outcomes in microcredit. 
\end{abstract}

\begin{document}

\clearpage

\maketitle

\section{Introduction} 

Personalized pricing, once restricted to the idealized construction of economic theory, is now squarely within the realm of possibility for firms newly equipped with a deluge of fine-grained information about individuals and prediction modeling of demand or willingness to pay based on this information. Given both the ubiquity of prices and their relevance in important domains such as hiring, lending, and credit subject to antidiscrimination regulation, price personalization remains an area of increasing scrutiny and caution, as well as tentative optimism due to competitive considerations \cite{house2015big,gonzaga2018}. 

The potential of expanded reliance on predictive models in domains affecting individuals is cause for concern. After all, the extensive study of fairness considerations in predictive models highlights how the joint structure of protected attributes and other information can lead algorithmic decisions, even based only on non-attribute information, to nonetheless lead to disparate impacts on individuals \cite{barocas-hardt-narayanan}. The setting of personalized pricing is particularly interesting because it fundamentally involves considerations of both resource allocation in response to price; as well as predictive models for such a price response. Auditing challenges arise precisely because valuations are in general not known or observed; rather only binary-feedback demand response is observed.

Studying the case of personalized pricing is \textit{conceptually} challenging because prices are a shared tool in drastically different domains: we consider lending/insurance, consumer goods, and public provision. A crucial distinction is between \textit{value-based} pricing that offers different prices to customers based on their estimated willingness to pay, and \textit{risk-based} pricing which offers different prices to customers based on their estimated costs, as in lending and insurance \cite{house2015big}. While discrimination law is strongest in insurance and lending, in lending, discrimination concerns often arise from individual agents providing offers from an actuarially-fair securitized rate sheet \cite{bartlett2019consumer}. In particular, distributional concerns regarding price optimization reflect overall concern for differentially adept/prepared/educated negotiating customers in insurance and lending, but slight optimism in value-based pricing since low-income individuals may be more price-sensitive \cite{bartlett2019consumer}. Hence, the majority of our analysis will focus on value-based pricing, which lends itself more readily to price optimization. 

In the case of value-based pricing, incidents where price targeting leads to disparities are often subject to media coverage and public outcry. We recollect just a few of these incidents: Staples changed prices based on available brick-and-mortar locations of competitors leading to higher prices for rural areas \cite{valentino2012websites} and Asians faced higher prices as a result of the Princeton Review's zip-code based price-targeting \cite{vafa2015price,larson2015unintended}. While these \textit{covariate-based} pricing schemes were based on non-group contextual information, nonetheless they induced disparities in prices along group lines. Although there are not clear anti-discrimination principles that govern the setting of value-based pricing, understanding the tradeoffs introduced by considering constraints or fairness penalties on a myopic price optimization problem can shed light on tradeoffs between various intuitive notions to inform algorithm design.

In this paper, we study the interplay of fairness and welfare considerations as they arise in the setting of personalized pricing. Our first task is conceptual: we square these real-world problem settings alongside previously expressed concerns regarding price optimization. For example, acknowledging the empirical reality of economic inequality informs the expected distributional impacts of covariate-conditional pricing schemes along racial and economic segments. We then turn to analytical modeling to identify the multiple objectives  which pricing decisions affect with fairness and/or welfare implications. Prices impose a burden on customers (perhaps only on those who purchase), result in allocations of the good itself, may be optimized based on noisy predictors, and the good itself may have downstream impacts of interest. In \Cref{sec-defs}, we taxonomize these considerations into price parity (marginal and conditional), model error fairness, preferences for access, allocative efficiency, and actuarial fairness; we offer these operationalizations (all of which may in themselves not be novel) alongside contextual discussion. 
We propose operationalizations of these normative considerations which avoid classical assumptions of known valuations, focusing on identifiability based on available information. In \Cref{sec-analysis} we provide further characterization of implications for fairness and welfare considerations for considering modifications to price optimization in order to improve on some of these notions, in particular price parity and market share. Where possible, we provide analytical insights on tradeoffs. In \Cref{sec-casestudies} we build empirical case studies on datasets related to the ``public-interest goods'' setting based on a study of willingness-to-pay for an elective vaccine and interest rates for microcredit loans.

\section{Related work}
The study of algorithmic pricing and revenue management is very extensive and spans economics, operations research \cite{gallego2019revenue}, and computer science. Price discrimination has also been studied for ethical and normative considerations, especially in relation to privacy \cite{odlyzko2004privacy,miller2014we}. We now highlight methodological and empirical work of particular relevance to algorithmic considerations. In \Cref{sec-conceptual argument} we discuss domain-level considerations and more broad related work in greater detail.

We first briefly overview the classical economic taxonomy of price discrimination \cite{varian1989price}. First-degree price discrimination offers individual prices to customers exactly at their willingness to pay, which is assumed to be known. Second-degree price discrimination depends on the \textit{quantity purchased} but does not differ across consumers, such as bulk discounts. Third-degree price discrimination charges different prices to different \textit{groups} of consumers, such as offering senior or student discounts. We focus on analyzing covariate-conditional prices, which are a form of third-degree pricing but draw nearer to first-degree pricing (up to the noise of random valuations).

The economic literature studies welfare (consumer and producer surpluses) implications of idealized first-degree personalized pricing, assuming valuation distributions are known. This classical notion of welfare is hence pegged to the valuation distribution (e.g., consumer welfare is valuation minus price). \cite{varian1985price} studies third-degree price discrimination; using first-order conditions of valuation distributions, they show that consumer welfare increases with additional price discrimination as long as total output increases. \cite{aguirre2010monopoly} provide analogous conditions for similar analysis. \cite{bergemann2015limits} shows that 
a seller can choose various segmentations that can achieve any combination of increase/decrease in consumer surplus; \cite{cummings2020algorithmic} study the theoretical computational efficiency of finding such segmentations.

 This classical theory suggests that ideal personalized pricing may improve welfare relative to uniform monopoly pricing, but that different segmentations lead to possibly indeterminate outcomes for consumer welfare. The empirical literature indicates this indeterminacy in important settings. \cite{fuster2018predictably} empirically study implications of machine learning predictors of default probability on disparities not only for predictive performance, but on using these predictions to set interest rates for loans via risk-based pricing. While richer machine learning predictors expand access to credit; they also result in greater price dispersion for the minority borrower on the margin. Hence, greater access comes at the cost of a greater price burden. \cite{dube2019personalized} study personalized pricing in a Bayesian setting with posterior uncertainty quantification in a business-to-business marketing setting; they find that finer-grained personalized pricing overall increases consumer welfare; though this is not monotonic in segmentation granularity. 
 
Pricing in the context of mechanism design follows another approach
 and assumes
  elicitation of ontologically valid valuations from strategic participants due to narrowly bracketed contexts such as auctions, kidney exchanges, and matching markets. \cite{dworczak2019redistribution} show that for a two-sided market, market-clearing competitive equilibrium pricing is not necessarily optimal if a market designer has distributional preferences. Noting that classical theory on quasi-linear utility ``implicitly embeds the assumption that each agent values money equally,'' they study implications of dispersion in marginal values for money of market participants for optimal market mechanisms.
 
\cite{cohen2019pricing} study fairness considerations when each protected group is assigned one price and the valuation distribution is known. In contrast, we focus on the prediction setting with rich covariates, and propose metrics that are completely independent of the valuation distribution. Our analytical insights focus on covariate-personalized prices and implications of joint structure of covariate distributions and group variability on fairness considerations.

 Finally, we mention work that studies tensions in fair machine learning, specifically the role of algorithms in allocating decisions or conferring utility, to highlight questions of interest that have analogies in the pricing setting. \cite{liu-drsh18,heidari2018fairness,hu2019fair,kasy2020fairness} broadly study tensions between fairness and welfare when machine learning enacts allocations, e.g., via classification. An interest of this work is to study analogous considerations for price optimization, under corresponding notions of fairness and welfare. 
 Longer-term considerations of fairness constraints have also been studied \cite{d2020fairness,creager2019causal}, typically with the formalism of dynamical systems.
 In machine learning, regulatory considerations barring ``disparate treatment,'' e.g., using the protected attribute information in the predictor or in algorithmic interventions, may be in fundamental tension with achieving proposed fairness notions. \cite{lipton2018does} studies tensions that arise from interventions that do not use attribute information. Again, these broad concerns may additionally be of interest in the setting of price optimization.

\section{Problem Setup} 

We let $X$ denote customer covariates (features) and $A$ the protected attribute. To simplify discussion we focus on binary comparisons between two protected groups, $A=a$ and $A=b$.  A personalized pricing policy is a function mapping covariates and possibly protected attribute information to a real-valued price, $p(X,A)\in \mathbb{R}^+$ (or, $p(X)$ for attribute-blind pricing rules). Each sale instance is associated with a hypothetical and unknown demand curve representing the demand for each possible price, $D(p)\in \mathbb{R}^+$. 
Often, when each sale instance is with an individual customer, $D(p)\in\{0,1\}$ is binary where $D(p)=1$ denotes the individual would purchase or take-up at price $p$. In \cref{itm:4}, we further restrict $D(p)=\indic{V\leq p}$ to be given by a customer valuation. Otherwise, we do not make this restriction, allowing arbitrary, possibly non-rational behavior.

We are primarily interested in studying the question of \textit{covariate-based pricing}. One approach is to estimate personalized demand at a price, given covariates, via a parametric or semi/non-parametric model, which we will denote as $D(p\mid x,a)=\E[D(p) \mid X=x]$.

A personalized pricing function satisfies: 
$$ \pxa \in \argmax_{p(x,a)} \E[ D(p(X,A)) p(X,A)] $$
Often, policy or domain-level restrictions may prohibit prices that directly use the attribute $A$: 
$$ \px \in \argmax_{p(x)} \E[ D(p(X)) p(X)] $$
We define $P=p(X,A)$ as the \textit{per-individual} personalized price; it may sometimes also refer to $P=p(X)$. We additionally introduce notation for the revenue, $\Rev(P) = PD(P)$, and its covariate-conditional counterpart.

We will also consider cases where the good has an effect on some outcome in itself, such as repeat purchasing behavior, health benefits, downstream welfare, so that $D(p)$ is also itself a \textit{treatment}. We denote the corresponding \textit{potential outcomes} as $Y(D(p))$. These represent the causal outcomes of take-up/non-take-up, e.g., the effect on contracting malaria of purchasing/not purchasing a bed net.

\section{Definitions and metrics}\label{sec-defs}
We now introduce definitions of different aspects of fairness or welfare in personalized pricing, addressing normative considerations motivated by different contexts, and offer formalisms and operationalizations of these. We consider a generic personalized price function $P$ which may reflect either first or third degree price discrimination.
In \Cref{sec-analysis} we discuss these operationalizations in more depth and analyze potential trade-offs.

\paragraph{Observational metrics.} 
We introduce the following notions of allocative fairness based on what we may \textit{observe}: prices $p$, demand outcome $D(p)$ (e.g., purchase/no purchase), and potential downstream outcomes due to the good $Y(D(p))$. 
In \Cref{sec-conceptual argument} we provide further discussion on why, if we are concerned about fairness in the first place, we might be skeptical about defining fairness relative to valuations or other \emph{un}revealed latent preferences/valuations. 

    \subsection{Price parity}\label{itm:1}
    
    \paragraph{Context.} A customer always benefits from a lower price. The difference between distributions of prices faced by different groups measures potential unfairness in price burdens. An extreme example is the so-called ``pink tax'': \cite{de2015cradle}, commissioned by the Mayor's office in New York City, studies gender-based pricing differentials and finds an average of 7\% higher prices paid by women; for example, women's pink razors are more expensive, for the exact same product\footnote{Notably, while legislation has been proposed to try to address the pink tax, this has not been established as gender-based discrimination. See \cite{jacobsen2017rolling} for more discussion.}. This highlights the capacity of price optimization to extract consumer value from behavioral failures of economic regarding valuations. In particular, using ``pinkness'' to segment products corresponds to extracting valuation from social constructions of gender, all other functionality being the same for a product with no signaling value. While the ``pink razor'' is an extreme example, considerations regarding price parity are a common intuitive objection to personalized pricing. 
    
    \paragraph{Operationalization.} 
    We introduce a definition based on distributional equivalence of prices for each group.
    \begin{definition}[Price parity]\label{def-price-parity}
$P \indep A $
    \end{definition}

We may also consider parity in moments of the price distribution, which also enables a simple way to give a scalar metric to disparity.
    \begin{definition}[Marginal price parity]\label{def-marginal-parity}
 $$\E[P \mid A=a] = \E[ P \mid A=b] $$
 Correspondingly, the marginal price \emph{dis}parity is $$\E[P \mid A=a] - \E[ P \mid A=b].$$
    \end{definition}
The notion of ``disparate treatment'' in fair machine learning suggests that the following notion of covariate-conditional price parity is intuitively appealing. 
\begin{definition}[$x$-conditional price parity]\label{def-conditional-parity} 
$p(x,a) = p(x,b)$
\end{definition}
Notice, however, that satisfying \cref{def-conditional-parity} generally does not ensure satisfying \cref{def-price-parity,def-marginal-parity} due to differing distributions of covariates between groups.
Indeed, in general contexts, it is well-understood that equal treatment need not lead to equal impact; this remains true in personalized pricing.

Finally, it is often helpful to consider price parity conditional on take-up, or more generally on demand. Conditioning on take-up reflects that price only affects consumer utility \emph{if} the customer purchases. 
\begin{definition}[Take-up-conditional parity]
$$  P \indep A \mid D(P) $$
\end{definition}
More generally, we may wish to condition on the \emph{effect} of pricing.
Consider the case where there is a nominal price $p_0$ and $P\leq p_0$ represents a personalized potential discount.
The event $D(P)>D(p_0)$ is the event that demand increases as an \emph{effect} of the discount. In the binary demand case: that an individual purchases if and only if given the discount (rather than purchasing irrespective of discount or not purchasing irrespective), which we term a \emph{responder} to the discount.
Conditioning on responsiveness accounts for the possibility that different groups have different valuations or willingness to pay, and to the extent that one deems it acceptable to personalize to leverage such differences (and often it is not) parity conditional on response requires we do not price-discriminate more than is justified by response to discount.
 \subsection{Model error fairness}\label{itm:2}

\paragraph{Context.} Given that fairness in machine learning studies how predictive models may exhibit differential model performance (predictive accuracy, error distributions) by group, a natural question is how such disparities in predictive model performance might affect the suboptimality of different prices; and whether different groups might experience different price suboptimality burdens due to error patterns of the predictive model.
\paragraph{Operationalization.} 
The rest of this paper studies pricing based on a true conditional demand model, $D(p \mid x,a)$. In practice however, only an estimate $\hat D(p\mid x,a)$ is available from observed data. 
For example, for the pricing problem: 
\begin{align*}
     \hat p^*(x) &\in \argmin \E[ \hat D(p(X)) p(X))  ]
\end{align*}
Price suboptimality fairness is concerned with the decision suboptimality of a price based on a risk model vs. the price derived from the actuarially fair ``true risk'', 
$\hat p^*(x) - \px$.
For example, $\hat p^*(x)$ may differ from $\px$ when we learn the prediction $\hat{D}(p \mid x,a)$ from finite samples and differential accuracy thereof could lead to fairness concerns.
In \Cref{sec-analysis} we will focus on the revenue objective.

 \subsection{Access and equal access}\label{itm:3}
 
 \paragraph{Context.} 
 We are often concerned about \emph{access} to the good being sold, especially when the good has benefits that are deemed crucial such as vaccines (see also our empirical study in \cref{sec: vaccine}), loans, or broadband internet.
 In terms of welfare, it is important to consider the total access that personalized schemes lead to, namely the total demand.
 In terms of fairness, we may be concerned with allocative parity in the form of parity in market shares or take-up probabilities by group. Personalized pricing schemes may in fact enhance both of these measures by allowing revenue extraction from high-valuation groups to enable offering lower price offers to low-valuation individuals, hence ``pricing people into the market.'' High-valuation groups are usually those with financial means that allows them to have a higher willingness to pay, and low-valuation groups are usually those with less financial means.

 \paragraph{Operationalization.} 
 Total access is simply the marginal demand. When demand is binary, this is the fraction of individuals who take-up the good.
 \begin{definition}[Total access]
 \begin{align*}\E[D(P)].\end{align*}
 \end{definition}
 The idea of access parity suggests requiring equal allocation of access / market share / take-up.
 \begin{definition}[Access parity]
 \begin{align*}\E[D(P)\mid A=a]=\E[D(P)\mid A=b].\end{align*}
 Correspondingly, access \emph{dis}parity is $\E[D(P)\mid A=a]-\E[D(P)\mid A=b]$.
 \end{definition}

 \begin{table*}[ht!]
	\begin{tabular}{lllll}
		&  Insurance & Lending & \thead{Consumer\\goods} & \thead{Public-interest \\goods} \\
\hline
				Moral hazard/Adverse selection &  \cmark         &  \cmark    &\xmark & \xmark  \\
Revenue-driven price optimization &\xmark &\xmark &\cmark &\cmark \\
Risk-driven price optimization & \cmark &\cmark &\xmark &\xmark \\
				\hline 
		Prefer marginal price parity &     \xmark     &   \xmark      &    \cmark               &              \xmark         \\
		Prefer conditional price parity&       \cmark    &  \cmark   & \xmark & \xmark                    \\
		Prefer access & \cmark & \cmark &\cmark & \cmark \\
		Actuarial fairness   &  \cmark         &  \cmark       &        \xmark           &        \xmark          \\
Allocative efficiency/sorting & \xmark & \xmark & \cmark & \cmark \\
Targeting long-run dynamics & \cmark & \cmark & \cmark & \cmark 
	\end{tabular}
	\caption{Different problem settings and \textit{what} fairness/welfare notions are relevant \textit{when}.}\label{tbl-settingscrossmetrics} 
\end{table*}

 \subsection{Allocative efficiency and fairness}\label{itm:4}
 \paragraph{Context.} In settings where interpreting revealed preferences as rational choices due to latent valuations, efficiency considerations are concerned with how prices \textit{sort} individuals by their valuations and ensure that a good may be \textit{targeted} towards those who value it the most.
 For example, the literature on pricing health interventions in development is particularly interested in the difference between free and low-price provisions based on whether prices better target households who are more likely to use a product  \cite{cohen2010free}. 
  An important further concern is whether errors in sorting individuals disproportionately affect one group more than another so that, on average, certain groups are more often incorrectly given priority over others.
 
 \paragraph{Operationalization. }
 We focus on providing observational metrics which assess sorting/targeting \textit{without} assuming access to the full valuation distribution. For this section, we focus on the binary feedback setting where $D(p) \in \{0,1\}$.

  \begin{assumption}[Monotonicity]\label{asn-monotonicity} 
  For any $p, p':$
$$ p > p' \implies D(p) \geq D(p')$$
\end{assumption}
Assuming monotonicity of binary demand with respect to price as in Asn.~\ref{asn-monotonicity} is equivalent to assuming a random latent threshold model, i.e., $D(p) = \mathbb{I}[ V \leq p]$. 
 This perspective recognizes that observations of the binary event $D(p)$, under Asn.~\ref{asn-monotonicity}, are censored observations of the underlying valuation.

 The question is whether a given pricing scheme (for which we have observed the binary outcomes) appropriately ranks valuations of individuals. A marginal measure of such efficiency may be \emph{concordance}:\begin{definition}[Concordance]Given two individuals drawn independently at random, \emph{concordance is} $$\pr(V_1 > V_2 \mid P_1 > P_2).$$
\end{definition}
While concordance captures how efficiently prices sort individuals by valuation, such efficiency may have disparate effects. To capture this disparity, we propose the \textit{class-crossed concordance disparity} metric.
\begin{definition}[Class-crossed concordance disparity]Given two individuals drawn independently at random from groups $A=a$ and $A=b$, respectively, \emph{class-crossed concordance disparity} is $$ \pr(V_b > V_a \mid P_a < P_b) -  \pr(V_a > V_b \mid P_b < P_a) $$
\end{definition}

 Class-crossed concordance has the following probabilistic interpretation. The term $\pr(V_b > V_a \mid P_a < P_b)$ can be interpreted as: of those whose valuations can be ordered under Asn.~\ref{asn-monotonicity}, what is the probability that valuations drawn from one group are \textit{stochastically greater} than valuations from another group? Class-crossed concordance measures a groupwise disparity in the difference in these probabilities.

\subsection{Targeting long-run dynamics}\label{itm:5}
\textit{Context.} A key domain-level consideration that justifies \textit{preferring take-up} is that take-up of the good is itself a treatment with a downstream outcome, such as future purchases/customer loyalty in e-commerce, net present value of continued borrowing \cite{karlan2019long}, or usage and downstream health outcomes of a preventive health intervention in development economics \cite{dupas2014short,cohen2010free,johnson2019}. Therefore, price impacts an allocation which itself may have heterogeneous effects on longer-term outcomes for the customer and/or decision-maker: we identify this as {targeting long-run dynamics}. This, for example, justifies an overall preference for expanding market shares. 

\textit{The possibility of a ``triple bottom line''.} Price personalization may be beneficial due to its increasing take-up of a good which is beneficial for individuals and the decision-maker, targeted price subsidies enable budget-balanced public provision (in contrast to a complete subsidy), and the access expansion might particularly those who would benefit the most. Of course, whether these benefits compound (or whether certain contributors are irrelevant) need to be assessed in the data of any particular setting, as in \cite{dupas2009matters,cohen2019pricing}.

\textit{Operationalization.} 
We recognize the firm's objective function as population welfare downstream of a price allocation: 
$$ \E[ Y(D(P)) ].$$

\subsection{Summary of problem settings and relevant notions}

Abstractly, we might summarize some of the above considerations by considering a $\lambda-$scalarized multi-objective optimization problem, which represents
the expansion of considerations beyond myopic revenue maximization: 
\begin{align}  \max_{P(\cdot)} &\;\; \E[P D(P)] 
\textstyle  + \lambda_1(\sum_{a \in \mathcal A} \E[D(P) \mid A=a] )  + \lambda_2 \E[Y(P)] \nonumber \\
\text{s.t. }& \E[ P \mid A=a] - \E[ P \mid A=b] \leq \Gamma \nonumber \\
& \E[ D(P) (P - c) ]\geq 0
\label{opt-problem-generic}
\end{align} 
\Cref{itm:1,itm:3,itm:5
} (price parity, access, long-run welfare) might conceivably be included in a conceptual ``multi-objective'' version of the firm's problem, while \Cref{itm:2,itm:4}
(price suboptimality, class-crossed concordance
) are idealized measures.

In \cref{tbl-settingscrossmetrics} we apply our conceptual taxonomy of problem domains to these different fairness/welfare notions in pricing. The first category of criteria summarizes how different problem settings differ in the presence of moral hazard/adverse selection that justifies risk-based pricing), and the capacity for price optimization. The second category identifies which notions of fairness or equity are more or less relevant in different settings. 

We caution that the above optimization problem is a conceptual device 
to illustrate how these notions might justify revenue sub-optimal allocations. \Cref{tbl-settingscrossmetrics} suggests that in any particular application setting, these notions may not be simultaneously relevant.

\section{Analysis of Definitions and Metrics}\label{sec-analysis}
In this section, we expand further on each definition. Where possible, e.g. by making additional assumptions, we provide analytical insight on implications of these fairness notions for price optimization or corresponding specializations of \cref{opt-problem-generic}.
\subsection{Price-parity}
We study the price optimization problem with additional price parity constraints, focusing on highlighting tradeoffs with implications for algorithm design.
To simplify the analysis, we make the following assumptions. We assume a partially linear demand model with a link function of the non-price, covariate-driven demand $\Dbarxa$ and a linear component for price elasticity. 
\begin{assumption}[Partially linear demand model]\label{asn-partially-lin-demand}
    $$D(p\mid x,a) = \beta_a p+ \Dbarxa.$$
\end{assumption}

\begin{assumption}[Downward sloping linear demand with respect to price.]\label{asn-downwardsloping-lineardemand}
	$\beta_a < 0, \forall a\in \Aa$
\end{assumption}
For example, the linear model corresponds to $\Dbarxa = \alpha+\gamma^\top x$. Linear demand is a common assumption for contextual pricing \cite{qiang2016dynamic,ban2020personalized,besbes2015surprising}. We also assume that price elasticities are negative.

Without loss of generality, assume group $a$ has higher average price at the unconstrained solution). The marginal parity\footnote{We discuss constraining first moments of the price distributions (marginal parity) to provide analytical insights. Constraining higher order moments via e.g. a set constrained by Kolmogorov-Smirnov statistic \cite{luo2020distributionally}, or a moment-based hierarchy as in \cite{aswani2019optimization}, is computationally possible.} 
 constrained revenue maximization problem, is a specialization of \cref{opt-problem-generic}. Let $\pstar(x,a;\Gamma)$ denote the corresponding $\Gamma-$parametrized solution.
\begin{align}
p^*(x,a; \Gamma) \in &\argmax_{p(\cdot)} \; \E[ p(X,A) D(p(X,A))] \nonumber \\
\text{s.t. } \; & \E[p(X,A) \mid A=a] - \E[ p(X,A) \mid A=b] \leq \Gamma \label{opt-marginal-parity} 
\end{align}
  The \textit{attribute-blind} personalized price is $\pstar(x;\Gamma)$, which restricts the above optimization to prices which only personalize on $x$. We derive the parity-constrained optimal price. 
\begin{theorem}\label{prop-price-equity} 
Let 
   $$\xi(A)= (\pr[A=a]^{-1} \mathbb{I}[A=a] - \pr[A=b]^{-1} \mathbb{I}[A=b]  ).$$ The optimal attribute-based personalized price under marginal parity solving \cref{opt-marginal-parity} is: 
\begin{align*} \pstar(x,a;\Gamma)  &= \frac{- \overline{D}(x,a) +  \xi(a) \lambda^*_{xa}}{2\beta_a},\\
  \text{where } \lambda^*_{xa}& = \frac{\E[\overline{D}(X,A) \xi(A)+ 2 \beta_A \Gamma]}{\E[ \xi(A)^2]}.
\end{align*}
The optimal attribute-blind personalized price is
\begin{align*} \pstar(x;\Gamma)& = \frac{- \overline{D}(x) + \lambda^*_x \E[ \xi(A) \mid X=x]  }{2\E[ \beta_A\mid X=x]},  \\
 \text{where } \lambda^*_x &=   \frac{\E[\overline{D}(X) \xi(A)  + 2 \beta_A \Gamma]}{\E[\E[ \xi(A) \mid X]^2]} .
\end{align*} 
\qed
\end{theorem}
The proof is included in \Cref{sec-proofs}; the key idea is to study the Lagrangian dual of the knapsack-constrained quadratic program and solve by swapping the order of minimum and maximum. 

In the following analysis, we focus on an equality constraint
in \cref{opt-marginal-parity}. Interpreting the solution, $\pxa$ differs from the unconstrained personalized price by a penalty whose size depends on the discrepancy of the price-independent covariate-based demand within groups.

We highlight some tradeoffs induced by marginal price parity against other fairness considerations. 
We first consider a very special setting where the demand function is invariant across groups; it is only the group-conditional covariate distribution which induces price disparity. 
    \begin{proposition}[Attribute-based vs. attribute-blind pricing under marginal parity ]\label{prop-price-equity-whowins}
    Suppose 
    Asns.~\ref{asn-partially-lin-demand},\ref{asn-downwardsloping-lineardemand}, and further that:
    \begin{enumerate}
        \item $\overline{D}(x,a) = \overline{D}(x,b)= \Dbarx$,
        \item $\Dbarx$ is linear in $x$,
        \item and $\beta_a = \beta_b$.
    \end{enumerate}
    Then we have \begin{align*}
        &  p^*(x,a;0) - \pstar(x;0) < 0 \iff\\
        & 
        \frac{ \lxa }{\lx}
< \pr(A=a\mid X=x) - \frac{\pr[A=a]}{\pr[A=b]}\pr(A=b\mid X=x),\\
  &  \text{ and }      p^*(x,b;0) - \pstar(x;0) < 0.
        \end{align*} \qedhere
    \end{proposition}       

In this very special case, we find that group $b$ would uniformly prefer attribute-based marginal parity pricing. The relationship is not necessarily uniform for group $a$. A sufficient condition, for example, is if for some $x,$
$\pr(A=a\mid X=x) >>\pr(A=b\mid X=x)
$
and $\pr[A=a]=\pr[A=b]$: then $\pstar(x,a;0)-\pstar(x;0)<0$. Hence, for $x-$outliers within group $a$ satisfying this condition, attribute-based pricing is Pareto optimal relative to attribute-blind pricing for \textit{both} groups. Since however price disparity in this special case is exactly driven by variability in $\pr(A=a\mid X=x),$ we might also expect that $\pstar(x,a;0)-\pstar(x;0) > 0$ for some other $x$. 

Building on the results of \Cref{prop-price-equity} and \Cref{prop-price-equity-whowins}, we also provide bounds on the revenue loss due to attribute-blind pricing, now in the setting where the non-price-based demand may differ across groups. 
\begin{corollary}[Revenue loss of $\px$]\label{cor-revenue-bounds}
Suppose Asn.~\ref{asn-downwardsloping-lineardemand}, $\beta_a = \beta_b$ and $\overline{D}(x,a) \neq \overline{D}(x,b)$. Then,
\begin{align*} 
&\E[ \Rev(p^*(X,A)) - \Rev(p^*(X)) ] 
\geq \frac{1}{4 \beta}  \E[\overline{D}^2(X)- \overline{D}^2(X,A)] 
\geq 0.
\end{align*} 
\end{corollary}

We now use these characterizations from \Cref{prop-price-equity}, \Cref{prop-price-equity-whowins}, and \Cref{cor-revenue-bounds} to study tradeoffs between price parity and other desiderata, in particular x-conditional parity, and summarize some implications for algorithm design. 
\begin{enumerate}
    \item A firm may generically prefer attribute-blind pricing to attribute-based pricing schemes due to regulatory considerations or x-conditional price parity (\cref{def-conditional-parity}).
    \item If achieving marginal parity is of interest in view of price disparities, attribute-blind marginal parity achieves lower firm revenue than attribute-based marginal parity. We provide a quantitative bound on the gap in a simple case in \Cref{cor-revenue-bounds}.
    \item Under the special case of \Cref{prop-price-equity-whowins}, for some $x$, attribute-based marginal parity is strictly preferable to attribute-blind marginal parity for \textit{both} groups.
\end{enumerate}
These considerations might outweigh an intuitive preference for attribute-blind pricing in the case of marginal parity.

\subsection{Price suboptimality fairness} 
We provide a decomposition that gives structural conditions on when the sign of the prediction error is informative of the sign of the mispricing or decision error. In particular, this highlights a distinction between analyzing fairness in data-driven optimal prices vs. fair prediction in machine learning.

\begin{proposition}[Price suboptimality error decomposition]\label{prop-pricing-error-decomposition}
Assume that $\nabla \hat D(\hat p^*(x)\mid x) \neq \nabla \hat D( p^*(x)\mid x)$. Up to first order terms,
\begin{align*}\label{eqn-price-suboptimality} 
  \hat p^*(x) - p^* (x)& =
  \frac{{\hat D(p^*(x)\mid x) - D(p^*(x)\mid x) }}{\nabla \hat D(\hp(x) \mid x)- \nabla \hat D(p^*(x)\mid x)} 
  \\
  &
  + p^* \left( 1+ \frac{\nabla \hat D(p^*(x)\mid x)- \nabla  D(p^*(x)\mid x)}{\nabla \hat D(\hp(x) \mid x)- \nabla \hat D(p^*(x)\mid x)} \right)
\end{align*} 
\end{proposition}\qedhere
The decomposition is not computable from observed data (since some quantities depend on the unknown $p^*(x)$). One implication is that the sign of $ (\hat p^*(x) - p^*(x)  ) $ ultimately depends on the sign of a few quantities: 
\begin{align*} 1&&\hat D(p^*(x)\mid x) - D(p^*(x)\mid x) && \text{estimation error}\\
2&& \nabla \hat D(p^*(x)\mid x)- \nabla D(p^*(x)\mid x) && \text{ gradient est. error} \\
3&& \nabla \hat D(\hat{p}^*(x)\mid x)- \nabla \hat D(p^*(x)\mid x) && \text{ price elast. subopt.} 
\end{align*}

There are some cases where we may be able to conclude the sign of decision error $\hat p^*(x) - p^*(x)$: we can conclude $\hat p^*(x) - p^*(x)  >0$ if $1,2,3$ are all positive. 

However, in general, the main implication of the above proposition is that the direction of \textit{decision disparity} is not immediate from \textit{prediction error} of $D(p(x)\mid x)$ alone: it also depends on estimation error of the gradient, and the difference in gradients due to suboptimality. It is more difficult to conclude implications of pricing decisions (and more broadly, optimization decisions) based on uncertain nuisance predictions.

Applying the above result to $\pxa - \px$ allows us to make a similar conclusion for the discrepancy of pricing with respect to attribute-based $D(p(x)\mid x,a)$ vs. the attribute-blind $D(p(x)\mid x)$ setting. 
\begin{align*}
   \textstyle &\pxa - \px =\frac{{ D(\px\mid x,a) - D(\px\mid x)}}{\nabla D(\pxa\mid x,a)- \nabla D(\px \mid x,a)}
   \\\textstyle &+p^* \left( 1+ \frac{\nabla D(\px\mid x,a)- \nabla D(\px \mid x) }{\nabla D(\pxa\mid x,a) - \nabla D(\px\mid x,a) } \right)
\end{align*}

\subsection{Market share}
We study a multi-objective version of \cref{opt-problem-generic} with additional weights on group-conditional market share objectives. We consider demand that arises from an underlying valuation distribution. (This is true  without loss of generality under Asn.~\ref{asn-monotonicity} of monotonicity.) The most general assumption that admits a concave price optimization program is \textit{log-concavity} of valuation distributions.
\begin{assumption}[Log-concave valuation distribution]\label{asn-logconcave}
    Suppose that $$V = g(x,a) + \epsilon,$$ where $\epsilon$ has a log-concave probability density function. 
\end{assumption}
We assume log-concavity so that the cdf and cumulative cdf of $\epsilon$ (effectively $D(p\mid x,a)$) are also log-concave. Log-concavity is quite general; log-concave pdfs include the normal, exponential, logistic, extreme value, Laplace, gamma, Weibull, etc.

Then the population-level market share personalized-price specialization of \cref{opt-problem-generic} is a concave program under Asn.~\ref{asn-logconcave}, since maximization is equivalent under the monotonic increasing log transformation: 
\begin{equation}\label{eqn-population-market-share} 
 \pstar(x,a;\lambda)
   \in \underset{p(\cdot)}{\argmax} \;
   \E[ (p(X,A)+\lambda) D(p(X,A))] 
\end{equation}
We may consider the \textit{attribute-blind} restriction of the above:
\begin{equation}\label{eqn-population-market-share} 
 \pstar(x;\lambda)
   \in \underset{p(\cdot)}{\argmax} \;
   \E[ (p(X)+\lambda) D(p(X))] 
\end{equation}
The above problems consider a \textit{population-level} market share penalty; we also consider group-conditional market share penalties. These may arise from the penalty formulation of the market-share constrained problem, where \textit{some} $\lambda$ is the optimal dual Lagrange multiplier for the constraints 
$
{\E[ D(p)\mid A=a]\geq \Gamma }
$.  
\begin{equation}\label{eqn-group-market-share} 
\max \{ \E[ p(X,A) D(p(X,A)) ] \colon \E[D(p(X,A)) \mid A=a ] \geq \Gamma_a \}
    \end{equation}
    
    Correspondingly, the \textit{attribute-based} group-conditional market share price $\pstar(x,a;\lambda_a)$ solves, pointwise over $x,a$: 
   \begin{equation} \pstar(x,a;\lambda_a) \in \underset{p(\cdot)}{\argmax}\; (p+\lambda_a/\rho_a) D(p\mid x,a), \;\;\forall x,a
   \label{eqn-group-market-share-pxa}
   \end{equation}

We study the sensitivity of the unconstrained-optimal attribute-blind and attribute-based personalized price to \textit{local increases} in the penalty parameter, $\lambda$, relative from the unconstrained optimal price, i.e. $\nabla_\lambda p^*(x,a;0).$ This describes {how much the price changes} in response to implementing distributional preferences for market share. Quantifying these sensitivities sheds light on the dependence on $R'',$ the second derivative of the (conditional) revenue function.
 \begin{lemma}[Optimality conditions for different penalties ]\label{lemma-opt-conditions}
Suppose Asn.~\ref{asn-logconcave}.
    \begin{enumerate}
        \item Population-level market share, attribute-based sensitivity is
 $$\textstyle \nabla_\lambda \pstar(x,a;0) =  {R''(\pstar(x,a;0)\mid x,a)^{-1}}{\pstar(x,a;0)^{-2}}.$$
    \item Population-level market share, attribute-blind sensitivity is $$\textstyle\nabla_\lambda p^*(x;0) =  {R''( \pstar(x;0)\mid x)^{-1}}{\pstar(x;0)^{-2}}.$$
    \item Group-level market share, attribute-based sensitivity
is $$\textstyle \nabla_\lambda p^*(x,a;0) = 
\frac{1}{\rho_a}
{R''(p^*(x,a;0)\mid x,a)^{-1}}{p^*(x,a;0)^{-2}} .$$

\end{enumerate}
    
    \end{lemma} 
    Observe that these sensitivities 
    are negative, under Asn.~\ref{asn-logconcave}. The proof, included in the appendix, identifies the (pointwise) optimality conditions of the constrained optimizations, \cref{eqn-population-market-share}, \cref{eqn-group-market-share-pxa}, and applies the implicit function theorem. 
    
    We highlight some implications of \Cref{lemma-opt-conditions} for algorithm design. 
\begin{enumerate}
        \item For larger $\abs{R''}$ (greater curvature), the less price decrease is required to increase market share, and conversely for smaller $\abs{R''}$, the larger price decrease is required. 
    \item Quantifying these sensitivities in terms of $R''$ highlights the \textit{revenue} implications of these price fairness changes. Considering a second-order expansion of the revenue, smaller $\abs{R''}$ suggests that the larger price decrease may not have extreme  \textit{revenue decrease}.
    \item Curvature also quantifies the rate of convergence of the optimal price, e.g. if optimizing over a parametrized pricing policy via M-estimation \cite{kosorok2007introduction}. Hence, finite-sample variability of the optimal price (which may be assessed empirically by bootstrapping) may suggest low curvature. This suggests a robust approach which ensures out-of-sample market share may incur small revenue tradeoff in the low-curvature regime. 
    \end{enumerate}

\subsection{Allocative efficiency: concordance}\label{sec-analysis-concordance}

Assumption \ref{asn-monotonicity}, of almost sure monotonicity, suggests that the combination of continuous treatment and binary outcome can be viewed as a censored observation of the valuation. Again, we do not assume access to the underlying realizations of valuation distribution, but study what may be concluded about valuations given that we only observe the censored realizations $D(P) = \indic{V>P}.$

Consider ranking the prices and valuations of two generic price-valuation-demand triples, $(p_1, v_1, D(p_1)), (p_2, v_2, D(p_2))$. The only joint outcome of demands and prices that admits concluding an ordering \textit{on the underlying valuations} $v_1, v_2$ is that $$\{p_1 < p_2, D(p_1) = 0, D(p_2) = 1 \} \iff \{ v_1< p_1 < p_2 < v_2  \}$$
Using this observation (which is highly dependent on almost sure monotonicity), we can identify a lower bound on concordance from observational data. 
\begin{theorem}\label{thm-concordance-equivalence}
Assume Asn.~\ref{asn-monotonicity} and $D(p) \in \{0,1\}$. 
$$ \pr(D_b(P_b) > D_a(P_a)  \mid P_a < P_b) 
\leq \pr(V_a > V_b \mid P_a < P_b)  $$
\end{theorem}

Note that $\pr(D_b(P_b) > D_a(P_a)  \mid P_a < P_b) $ is related to the concordance index of sensitivity analysis, in particular the perspective studied by \cite{steck2008ranking} 
who suggest a ranking-based approach to survival analysis. 

In survival analysis, right-censoring occurs when there is a finite horizon end to data collection for the survival time of patients, so that the observed survival times are the minimum of the censoring time and the actual survival time. The concordance score is a generalization of Wilcoxon-Mann-Whitney statistics and the AUC that applies to continuous output variables, and accounts for censoring of the data. It is the fraction of all pairs of subjects whose predicted survival times are correctly ordered among all subjects that can be ordered. \cite{steck2008ranking} observe that two subjects' survival times can be ordered not only if both of them are uncensored; but also if the uncensored time of one is smaller than the censored time of another. 

Relative to the concordance index of right-censored survival analysis, the setting of allocative efficiency is more difficult: we are required to further restrict attention to pairs $p_1 < p_2$, and we can at best order $v_1< p_1 < p_2 < v_2$.

\subsection{Targeting for long-run dynamics: optimal encouragement designs} 

An important justification for cross-subsidy (preferring take-up) is in recognizing that take-up of the good, $D(p)$ may itself be a treatment for downstream outcomes. Such outcomes of interest might include long-term customer value, social learning, repeated purchase behavior, ``compliance'', attrition, etc.  We highlight that one might view $p$ as either a continuous treatment or instrument. 

For example, this is a major focus of \cite{karlan2019long} which considers the amortized net present value of customers over a long time-horizon after they take-up a microcredit loan. Their analysis suggests that while price discrimination to expand take-up may result in losses in short-term profits, this can be outweighed by clients ``aging into'' a loan portfolio and becoming more profitable. This is also of concern for health interventions in development: take-up of the good is not the final outcome of interest, but rather health outcomes are. 

In particular, the possibility of such a ``triple bottom line'' highlights a situation where \textit{non-ideal theory} that propagates the effects of known inequality to the expected failures of classical economic theory may highlights possible opportunities for personalized pricing to achieve practical benefit. For example, a plausible narrative recognizes that poorer households ``undervalue'' preventive healthcare not because of some fundamental ``underlying preference'' for poor health, but due to behavioral considerations and cognitive burdens which prevent endogenizing the full health benefits of a product \cite{bar2018algorithmic}, or for far more practical reasons since they may simply have lower incomes. As a result, they may be more price sensitive. And, they may receive outsized additional health benefits from using the preventive health intervention if they are indeed liquidity constrained due to lack of other ancillary health interventions. Personalized price offers allow making lower price offers, increasing take-up, and if indeed these households ``priced in on the margin'' receive greater heterogeneous benefits, larger welfare improvements. 

There are two perspectives. In \Cref{fig-price-as-instrument}, price is a {continuous instrument} for treatment, and hence, outcome\footnote{Estimating a covariate-conditional local instrumental variable curve, or optimal policy when the price instrument is the control variable, remains an open problem. See \cite{kennedy2019robust} for doubly robust estimation of the continuous instrument or \cite{syrgkanis2019machine} for policy optimization with heterogeneous effects and discrete instruments.}. That is, recognizing that one cannot directly assign the actual treatment of interest -- one can neither force loans upon individuals nor ethically randomly reject individuals who apply for loans, exogenous price variation may be the only tool from observational data for assessing the causal impacts of loans on welfare. Crucially, price satisfies the main assumptions for instrumental variables; \textit{relevance}, that it predicts treatment (loan take-up), and more importantly the \textit{exclusion restriction}: that $Y(D(p)) = Y(D), \forall p,$ the only  effect of the price on outcome is via its impact on treatment \cite{imbens2015causal}. This may be plausible when the outcome of interest is a quantity such as health impacts of a bednet on malaria incidence \cite{bhattacharya2012inferring}, impacts of sanitation on health outcomes,  \cite{johnson2019}, or social learning for sustained use of the intervention via subsidized first use \cite{dupas2014short}. In the lending setting, this may be plausible if it is believed interest rates do not affect default event, or the amount borrowed. 

Another perspective views {price as continuous treatment}, e.g. \Cref{fig-price-as-treatment}. There are some posited behavioral economics effects which may lead to a failure of the exclusion restriction such that price affects outcome, such as anchoring to reference prices (which attenuates future take-up) or sunk-cost fallacies (when high prices  encourage usage/non-wastage); this is explored in \cite{dupas2014short}. Alternatively, interest rates might affect default probability if individuals are liquidity-constrained. Evidence is mixed in lending: \cite{fuster2018predictably} assumes this, \cite{alan2013subprime} finds no effect, and \cite{karlan2008credit} finds some effect of rates on default. Alternatively, interest rates might have an effect on the extensive demand margin (amount borrowed). In this setting, we might instead consider price as a continuous treatment with a composite outcome of take-up and observed outcome, conditional on take-up (such as amount borrowed or default outcome). 

From the perspective of optimization, we generally view price as a treatment and optimize for corresponding downstream outcomes, e.g. conduct an ``intention to treat'' analysis.

\begin{figure}
\centering
\begin{subfigure}{0.25\textwidth} \centering
\begin{tikzpicture}
    \node[state] (X) {$X$};
    \node[state] (P) [above =of X] {$P$};
    \node[state] (T) [right =of X] {$D$};
    \node[state] (Y) [above  =of T] {$Y$};

    \path (X) edge node[above] {} (P);
    \path (P) edge node[above] {} (T);
    \path (P) edge node[above] {} (Y);
    \path (X) edge node[above] {} (T);
    \path (X) edge node[above] {} (Y);
\end{tikzpicture}
\caption{Price as treatment}\label{fig-price-as-treatment} 

\end{subfigure}\begin{subfigure}{0.25\textwidth} 
\centering
\begin{tikzpicture}
    \node[state] (X) {$X$};
    \node[state] (P) [above =of X] {$P$};
    \node[state] (T) [right =of X] {$D$};
    \node[state] (Y) [above  =of T] {$Y$};

    \path (X) edge node[above] {} (P);
    \path (P) edge node[above] {} (T);
    \path (X) edge node[above] {} (T);
        \path (T) edge node[above] {} (Y);
    \path (X) edge node[above] {} (Y);
\end{tikzpicture}
\caption{Price as instrument}\label{fig-price-as-instrument} 
\end{subfigure}
\end{figure} 
\section{Case Studies}\label{sec-casestudies}

\subsection{Willingness to pay for elective vaccine}\label{sec: vaccine}

We build a case study from \cite{slunge2015willingness}, a willingness-to-pay study for vaccination against tick-borne encephalitis in Sweden. The vaccine for tick-borne encephalitis (TBE) is elective and the study is interested in assessing determinants of willingness to pay to inform health policy. Demand is associated with price and income; as well as individual contextual factors such as age, geographic risk factors, trust, perceptions and knowledge about tick-borne disease. The health policy recommendation uses the learned demand model to estimate the vaccination rate under a free, completely subsidized vaccine. This setting corresponds to the setting of public provision,
where a decision-maker has a preference for higher take-up due to dynamic externalities of vaccines (which are nonetheless difficult to precisely estimate or target). The study was a contingent valuation study which asked individuals about take-up at a random price of 100, 250, 500, 750, or 1000 SEK. The study finds that ``The current market price of the TBE vaccine deters a substantial share of at-risk people with low incomes from getting vaccinated.''

\begin{figure*}[ht!]
    \centering

    \begin{subfigure}[b]{0.25\textwidth}
    \centering
    \includegraphics[width=\textwidth]{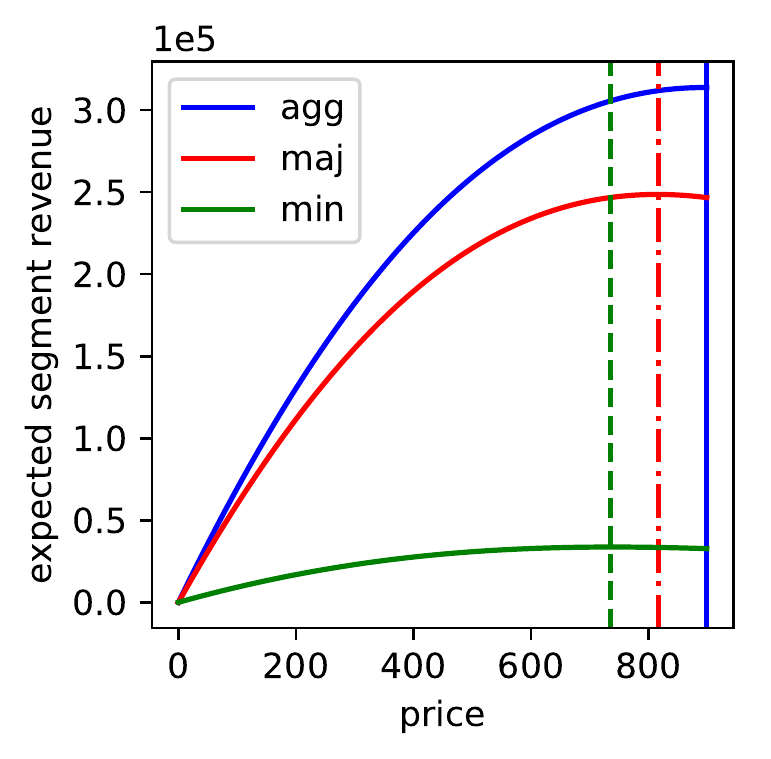}
    \caption{Revenue curves for \\$A= \{a,b\}, a, b$ segments}\label{fig-vaccinewtp-revcurves}
    \end{subfigure}\begin{subfigure}[b]{0.25\textwidth}
        \centering
    \includegraphics[width=\textwidth]{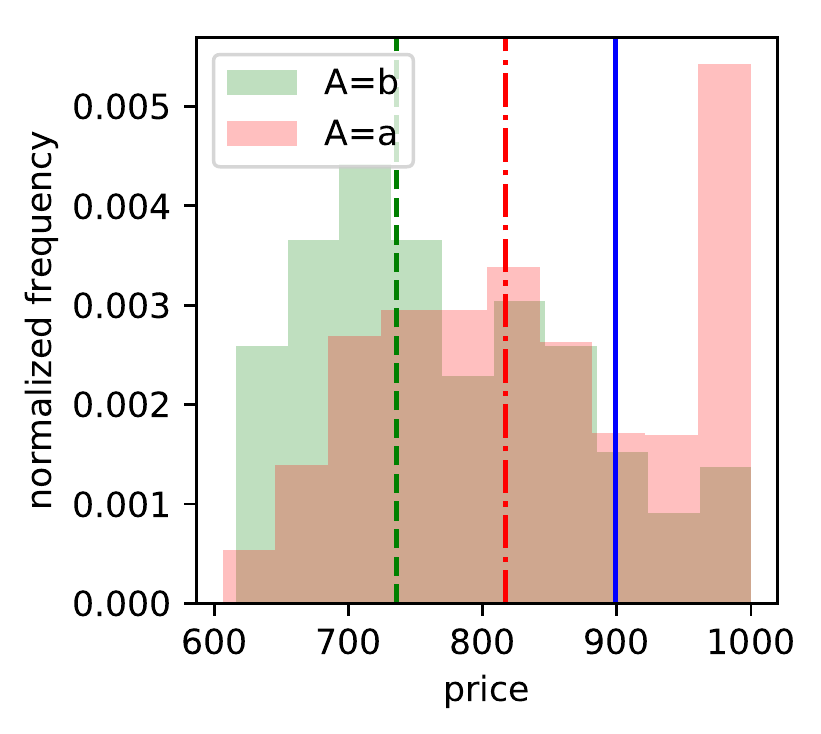}
        \caption{Distribution of \\$p^*(X)\mid A=a, p^*(X)\mid A=b$ }\label{fig-vaccinewtp-pstarX}
    \end{subfigure}\begin{subfigure}[b]{0.25\textwidth}
    \centering
\includegraphics[width=\textwidth]{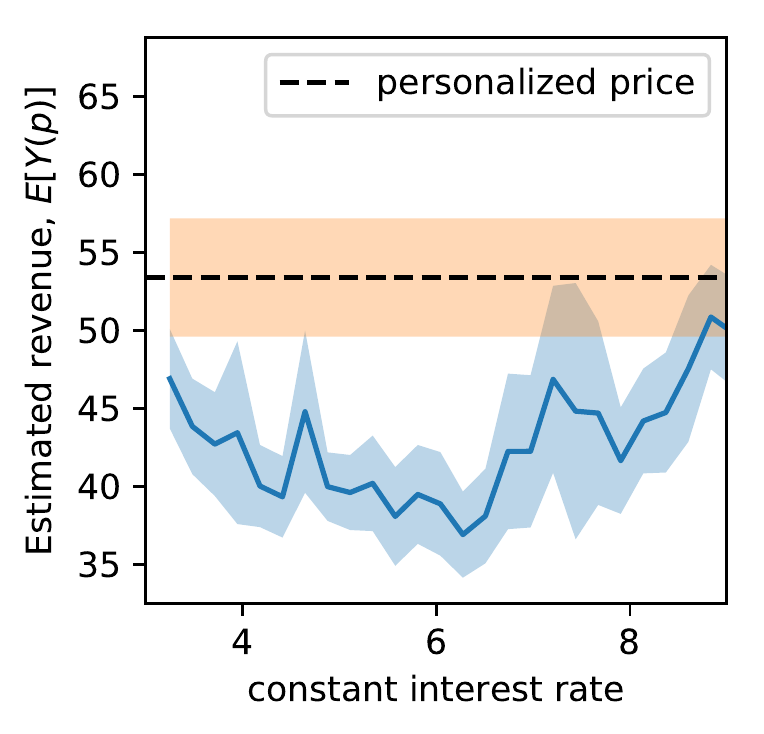}
\caption{Estimated revenue \\comparison}\label{fig-rev-comparison}
    \end{subfigure}\begin{subfigure}[b]{0.37\textwidth}
    \centering
\includegraphics[width=\textwidth]{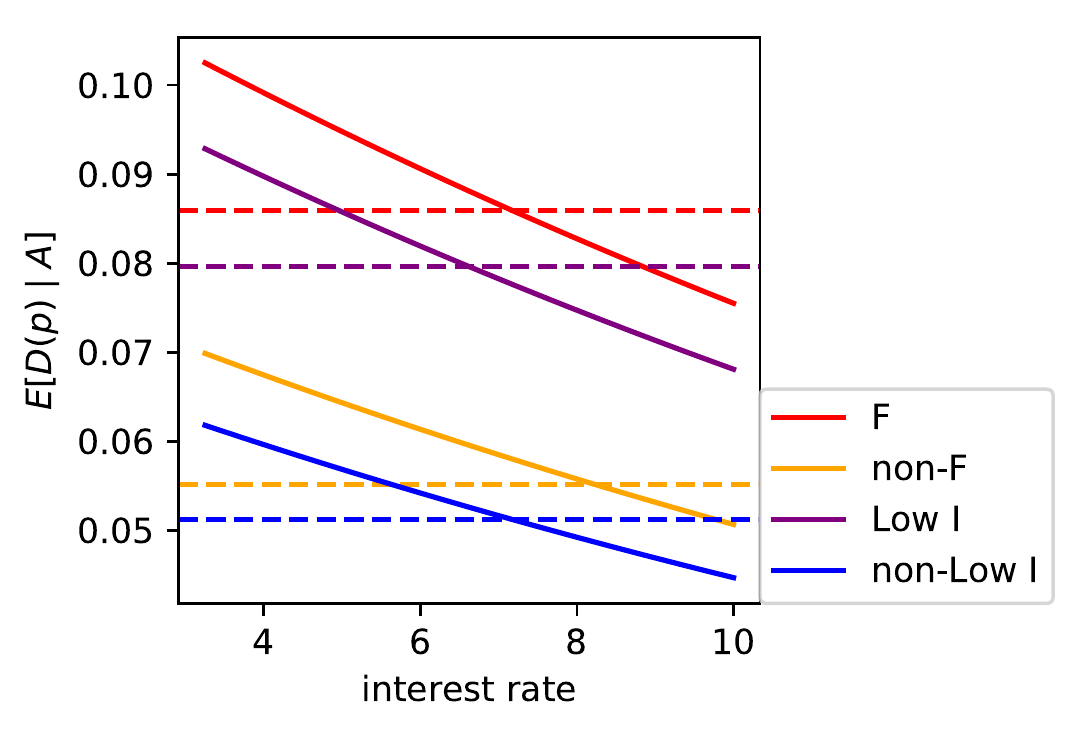}
\caption{Comparing take-up}\label{fig-takeup-microcredit}
    \end{subfigure}  
    \caption{Willingness to pay for elective vaccine (\Cref{fig-vaccinewtp-pstarX,fig-vaccinewtp-revcurves}) and Microcredit (\Cref{fig-takeup-microcredit,fig-rev-comparison}). }
    \label{fig:my_label}
\end{figure*}

In \Cref{fig-vaccinewtp-revcurves,fig-vaccinewtp-pstarX}, we compare distributional considerations of segmented vs personalized pricing. Let $A=b$ indicate low-income. We follow \cite{slunge2015willingness} and learn a logistic regression model of binary demand by simply appending the price covariate with the other covariates, so that $\Dmnd(x,p) = \sigma(\gamma^\top x + \beta p)$. A natural approach in the setting where a free subsidy is not feasible due to budget constraints,
 is \textit{third-degree} price discrimination:
 segment based on income and offer a price to low-income and high-income groups separately.
 We consider such a group-segmented approach in \Cref{fig-vaccinewtp-revcurves}. The blue curve (``agg'') represents the revenue curve for a uniform price. The red (``maj'') and green (``min'') curves are revenue curves from the majority (high-income) and minority (low-income) groups, respectively.
 We indicate the resulting optimal prices for each of these curves with vertical dashed lines. Notably, the $A=a$ revenue curve has greater second-degree curvature than the $A=b$ revenue curve. Because of the flat revenue curve, the market share of the minority group can be substantially increased without much extra cost. 

We next consider a covariate-driven personalization approach. In \Cref{fig-vaccinewtp-pstarX}, we plot histograms of the group-conditional distributions of these optimal prices, ${p^*(X) \mid A=a}$ and ${p^*(X)\mid A=b}$. The optimal group-based prices are indicated by the vertical lines for reference. We solve $$\px \in \underset{p(\cdot)}{\argmax} \;\; \E[D(p(X))p].$$ Notably, the optimal prices for the low-income group are overall lower than those for the high-income group. Expected take-up in this segment increases to $27.5\%$ from $24.2\%$ under the uniform monopoly price or $26.7\%$ under $p^*(b)$, the optimal group-based price of \Cref{fig-vaccinewtp-revcurves}. Compared to the uniform monopoly price which obtains expected optimal revenue of $313.6\cdot 10^3$, the group-based segment scheme $p^*(A)$ obtains expected revenue of $282.2\cdot 10^3$, and the personalized pricing scheme $p^*(X)$ obtains expected revenue of $318.8\cdot 10^3$. In this setting, covariate-based personalized pricing is strictly beneficial in terms of {(mild) \textit{revenue benefits} that are also able to achieve \textit{greater market share} for the minority group}. While the group-based segmentation results in a lower price for the minority group, it is overall not incentive-compatible for a decision-maker to use this segmentation because it attains less revenue than even uniform monopoly pricing.

\subsection{Credit Elasticities}
  \cite{karlan2008credit} randomize prices for a microfinance lender for repeat borrowers. An extensive literature on microfinance sought to assess whether microcredit was able to provide longer term benefits in improving outcomes for household. The rise of the sector led to partially subsidized lenders as well as interest from the private sector. The question of the study was to leverage price randomization in the microcredit setting and assess the effects of lower, or higher, interest rates on revenue for the lender. Overalal, the findings suggest lower rates could decrease profits by a small amount. But the paper considers that at the domain level, since microfinance initiatives may have targeting preferences, e.g. for financial inclusion for women or lower-income individuals, such potential mild profit losses could be offset by expanded inclusion of these target groups due to heterogeneity in take-up. 
  
  This setting could present an opportunity for personalized pricing to differentially lower interest rates and expand revenue to targeted groups. Adopting an ``intention to treat'' analysis, we use the method of \cite{kallus2018policy} to consider off-policy evaluation and optimization of a continuous linear personalized pricing policy from the randomized controlled trial data. The policy parametrization is linear in the covariates, which include income, demographics, location, and loan history information. The method of \cite{kallus2018policy} considers a kernel-based estimator of the counterfactual value of a pricing policy. We use the Epanechnikov kernel and a bandwidth of $0.3$; the optimization is non-convex. Because of the fundamental problem of causal inference, we lack the ability to directly assess outcomes.

Nonetheless, we provide some comparison of the estimated revenue and market shares under the personalized policy. We consider a 50/25/25 training/nuisance estimation/validation split, training a random forest on the nuisance estimation split, and learning an optimal policy on the training data with a doubly robust estimator. We use the random forest to estimate the revenue of the personalized policy in comparison to constant interest rates on the validation set.\footnote{This is in general a biased ``direct method'', and we take care to avoid extrapolation from interest rates. However, for the sake of comparing against constant treatment assignment, using the direct method reduces variance.} Finally, to indicate the sampling variation in our comparison induced by training the benchmark model, we repeat draws of the nuisance/validation sets, and report the sampling variation in revenue estimates via confidence bands of one standard error. 

We include the results in \Cref{fig-rev-comparison} and \Cref{fig-takeup-microcredit}. \Cref{fig-rev-comparison} plots the random-forest imputed revenue of the personalized allocation (indicated in black dashed, plotting one standard error), comparing against the imputed revenue via the random forest model of constant interest rates (on the x-axis), in blue. The personalized allocation rule increases estimated revenue (as expected). We assess some of the distributional characteristics of the resulting allocation. In \Cref{fig-takeup-microcredit}, we compare the access properties of the personalized decision rules for subgroups of interest, namely female and non-female borrowers and low-income and non-low-income borrowers. The estimates of access in \Cref{fig-takeup-microcredit} are based on a logistic regression of demand, learned on the nuisance estimation dataset.

In horizontal dashed lines, we plot the access estimates for these subgroups under the personalized allocation rule. Note the achievable subgroup access levels correspond to intersections of vertical lines with the demand curves. In comparison, the personalized pricing allocation rule is able to increase access for female and low-income borrowers, relative to the optimal constant interest rate (around 7.2). The unconstrained optimal personalized price however achieves lower takeup for non-low income borrowers (i.e. it increased interest rates for them). Stronger distributional guarantees may be possible by further constraining the price optimization problem. Overall, the goal is to highlight that personalized pricing can improve firm revenue, as well as increase access and improving targeting abilities.

 \bibliography{fair-pricing,beyondfairness}
\bibliographystyle{abbrvnat}
 \clearpage
 \appendix
 \onecolumn
 
 \paragraph{Overview of the appendix} 
 \begin{itemize}
     \item \Cref{sec-conceptual argument} provides a literature review of how considerations regarding price personalization have arisen in different problem settings. 
     \item \Cref{sec-proofs} provides proofs of the analysis in \Cref{sec-analysis}.
     \item \Cref{sec-data-details} provides further details on the empirics.
 \end{itemize}
 
 \section{Literature review }\label{sec-conceptual argument}
\subsection{Different problem settings (summary) }

\paragraph{Lending} Discrimination in lending is an important problem subject to regulation by the CFPB. Not only do banks decide whether or not to offer loans or extend credit (previously considered as classification problems), they also decide on the terms, i.e. interest rates, of the loan or credit, using risk-based pricing. These rates have significant welfare implications for individuals if discrimination leads to differences in rate terms. Standard discretionary pricing practices, where individual loan agents have some discretion to set rates above/below securitized rates based on local operating costs and competitive factors, may lead to disparate impacts due to, e.g. differential consumer bargaining leverage or discrimination. 

There is extensive literature studying household finance, including credit constraints of subprime borrowers with heterogeneity. For example, firms may find via price optimization that they are able to raise prices on subprime borrowers without affecting demand for credit-constrained consumers. 

Price optimization in the lending setting is primarily based on discretionary markups/discounts, e.g. negotiated on an individual basis with consumers. There are subsets of loans that are securitized by the government, which are presumed to be actuarially fair prices. Price dispersion beyond this can be explained a combination of individual discretionary pricing (negotiation) or market concentration (price optimization). A common strategy in papers that study potential discriminatory pricing in lending (e.g. \cite{bartlett2019consumer}) is to regress deviations from the Fannie Mae-securitized rate schedule on controls (e.g. for location-varying costs of lending); the coefficient on race in such a regression corresponds to unexplained disparate impact in pricing.   
\paragraph{Insurance} Insurance is the one of the originating domains of risk-assessment. However; it has the least available data. Our analysis correspondingly does not focus on or shed light on important questions in this area. 
\paragraph{Consumer products} There are growing concerns regarding the use of price optimization for consumer goods. Many concerns in consumer product pricing are complicated due to the difficulty of disambiguating consumer valuation based on ``legitimate'' aspects such as individual preferences (which may be culturally conditioned; hence associated with categories) vs. aspects that may seem ``illegitimate'' or repugnant to extract rent from consumers on the basis of. A different source of concern arises when universal provision of a good is expected.

However, studies show that consumers react strongly negatively to perceptions of price unfairness (perception is an important mediator, because different formats for the same posted price tend to have different fairness perceptions). This consumer backlash may be so strong it is posited as a reason that retailers do not wish to conduct extremely fine first-degree price discrimination. Therefore, we might be interested in general notions of price equity. 

\paragraph{Goods with public externalities} 

In this setting which arises in development, public, or health economics, a centralized decision-maker (DM) has some utility or preference for individuals receiving the good in addition to individuals' idiosyncratic distributions of willingness to pay. Price optimization is beneficial because it can subsidize participation in the market and take-up of the good by pricing at willingness-to-pay for individuals with high valuations to subsidize lower price offers to lower willingness-to-pay \cite{cohen2010free,dupas2009matters,johnson2019}. Cross-subsidy is particularly beneficial when understanding willingness-to-pay is in part related to ability to pay, which is commonly discussed in development economics with regards to credit-constrained consumers.

\subsection{Implications of price optimization in lending}

	\paragraph{Lending and implications of regulation} 
	In this setting, such as risk-based pricing in insurance and discretionary pricing in mortgage lending, discretionary pricing is highly regulated and the leeway for discretionary pricing or ``price optimization'' (rent extraction from consumers) is severely limited. Limited discretion is given to lending agents, for example to match competing offers or possibly (under interpretations of fair lending law) to cover operating costs (for example in certain geographic regions), e.g. ``justified business necessity''. \cite{bartlett2019consumer} study mortgage lending for loans bought by Fannie Mae, which are subject to price schedules based on creditworthiness, and find that while financial tech companies that engage in algorithmic pricing reduce discrimination on LatinX and African-American borrowers by 40\%, nonetheless bias still persists (as measured by residual coefficients on race in a regression of interest rates on controls).
	
	 Nonetheless, concerns about actuarial risk may still arise when actuarial risk is estimated from covariates. 
		\cite{bartlett2019consumer} summarizes their interpretation of discrimination law and its implications for the legitimacy of personalized pricing: 
	\begin{quote}
		(a) Scoring or pricing loans explicitly on credit-risk macro-fundamental variables is legitimate; (b) Scoring or pricing on a proxy variable that only correlates with race or ethnicity through hidden fundamental variables is legitimate; (c) Scoring or pricing on a proxy variable that has significant residual correlation with race or ethnicity after orthogonalizing with respect to hidden fundamental credit-risk variables is illegitimate.
	\end{quote}

\paragraph{Distributional concerns of interest in insurance/lending and risk-based pricing.}
In a recent case ``Association belge des Consommateurs Test-Achats ASBL and Others v. Conseil des ministres" (\cite{euinsurance}), the EU banned the use of sex for pricing in insurance policies. \cite{avraham2013understanding} points out some aspects in which the redistributive transfers which result may be preferred to transfers implemented through the typical public economics perspective of tax-and-transfer because the magnitude is pegged to the difference in marginal risk by group; and the transfer does not distort labor incentives as do changes in tax schedules. 

\cite{bartlett2019consumer} points out that regulators may have distributional concerns about price discretion in lending that differ from other settings (such as revenue management and pricing). In revenue management and pricing, and more generally settings with high fixed costs and lower marginal costs, price inelastic customers such as business-class travelers may subsidize access to a service via fare or price discrimination. In contrast, price-sensitive consumers in insurance and lending may be more fiscally savvy and/or ``shop around'' more for better terms; there is some concern that underprivileged borrowers may also be less equipped to leverage competition as such. This points to the importance of using \textit{non-ideal} theory and domain-specific considerations in motivating considerations of fairness. 

\paragraph{Underlying normative considerations barring risk classification based on protected attribute.}

\cite{avraham2013understanding} points out that normative grounds for arguments against risk classification based on protected attribute include luck egalitarian arguments (not holding individuals responsible for factors out of their control, which is difficult with risk to operationalize), as well as the ontological invalidity of social classifications for absolute risk except insofar as social categories designate the effects of inequities in history. Additionally, the consequences of inequity in accuracy of risk classification, arguments regarding causality (which however are more effectively levied along the lines of differential accuracy due to the presence of unobserved confounders) and privacy.

\paragraph{Concerns about price optimization: rent extraction from behavioral bias}

A key concern in household finance is in explaining empirical puzzles that contradict predictions from conventional economic theory. To this end, behavioral household finance considers implications of behavioral economics for explaining some of these puzzles. 

However, the expansion of the consumer credit market in the 1970s and 1980s led to concerns regarding rising debt, and many puzzles in household finance where empirical phenomena contradicted the assumptions of classical economic theory. \cite{block2005myth} studies the ``myth of the rational borrower'' in the context of the discussion about bankruptcy law. The central controversy was about whether or not the law provides incentives to declare bankruptcy, or if increases in the number of bankruptcies are instead driven by circumstances typically out of a borrower's control such as income shocks. They argue that a key counter-argument is ``the myth of the rational borrower''; that behavioral biases from economic theory lead suboptimally rational borrowers to overborrow relative to their actual marginal returns to credit. Whether or not consumers are behaviorally biased, or suboptimal in ways predicted by behavioral economics, is a research topic: \cite{wright2006behavioral} counter-argues that empirical evidence is not so strong that deviations from predictions of economic theory are explained by behavioral economics. Behavioral household finance remains an active area of study; although a meta-analysis on the evidence for the effectiveness of financial education interventions suggests that they are overall ineffective \cite{fernandes2014financial}. We refer to \cite{beshears2018behavioral} for a fuller discussion.

We argue that the main takeaways to guide our analysis are therefore: 
\begin{itemize}
	\item Conventional pricing theory and welfare analysis that assumes revealed purchasing behavior reflects valuation may be inappropriate under strong evidence of behaviorally suboptimal consumers, and in particular concerns about credit- and liquidity-constrained customers, especially poorer households. 
	\item Price optimization based on behavioral considerations may be profit-maximizing, but value-based pricing in this setting may be suspect (or may introduce distributional concerns opposite the slight favor in revenue management for price optimization). Competitive considerations may support the utility of price optimization, but we will abstract away from most competitive considerations and focus on implications for an idealized actuarially-fair pricing problem.
\end{itemize}

\subsection{Perceptions of unfairness from consumers due to different prices (equal treatment)}

It is unclear whether there is sufficient regulation to merit equal treatment as a normative rule for pricing, or to understand this squarely within the sphere of discrimination law. Some consumer research such as \cite{haws2006dynamic} studies consumer perceptions of fairness. Since in fact consumer reaction may differ to personalized pricing based on \textit{salience} -- e.g. showing prices vs. showing discounts -- the inconsistency of these reactions themselves suggests that intuitive perceptions of fairness in pricing do not reflect rational economic behavior; but rather that behavioral considerations interact with the predictions of standard expected utility frameworks. 

There are many high-profile instances where differential pricing appears exploitative but we 
would argue that what is doing the normative work is more broadly related to concerns
regarding perpetuating historical injustices or overt extraction of customer rent for reasons beyond their agency and correlated with historical inequity. In that light, the normative force is really in the "leakage" of structural considerations and the correlation between price optimization and these specific aspects. Conversely, it is difficult to reason abstractly about these settings without further grounded context. 

We posit some examples. Consider the razor tax. The razor tax is objectionable because it extracts customer welfare on the grounds of gender roles for functionally the same product. At some level, our objection is far more difficult to levy on the level of \textit{homophily or taste} associated with ``pinkness'', than it would be to levy on the objection that societal expectations regarding grooming have led to endogenizing higher willingness to pay. Another example is 
higher prices for shipping/delivery to certain zip codes (or not providing service at all: Here, there is some expectation of equitable service provision, or not reifying historical disadvantages due to location; even though antitrust provisions would allow charging different prices based on differential costs of doing business.  
Another important point contrasts price optimization based on first party vs. third-party data. Here, there is a violation of privacy norms. Consumers may object to certain kinds of data being used for price optimization vs. other kinds of data.

Disentangling aspects of valuations driven by the social constructions of race and gender, e.g. via mediation analysis or other counterfactual notions, is likely a very difficult task to operationalize. 

\subsection{Personalized pricing and public provision}

In this specialized setting, price optimization allows significant benefit for cross-subsidizing allocation of a good. Cross-subsidization is particularly beneficial when a central decision-maker has a utility preference for more individuals obtaining the good vs. not; for example in situations with positive externalities. 

These pricing schemes are of particular relevance for provision of health interventions in development, where targeting can ensure take-up of households of distributional interest \cite{johnson2019}. A natural question is, why not provide the good for free? The answer is due to a combination of {scarce resources} and efficacy that requires {take-up/compliance}. In these scarce-resource settings in health and development, if free provision to all were possible and feasible, that would be optimal. However, when resource constraints are binding, and the effects of health interventions are also realized only by ``compliance'' of the consumer in using it (preventive healthcare, bednets, contraceptives), the DM prefers subsidies to increase access to those who would use the intervention if offered. So there are additional targeting concerns that prefer allocation to individuals who would take-up and use the health intervention; that is, the DM is interested in balancing over-provision and over-exclusion \cite{cohen2015price}. To this end, price discrimination is considered a tool to screen-in individuals who would use the good, rather than simply allocating for free.

This setting is common in development and public economics. \cite{johnson2019} conceive a pricing mechanism to price individuals into a market for mechanical desludging in Kenya; a health intervention that has positive externalities but which may be out of reach for the poorest households. Estimating a demand and reservation price model, they learn personalized prices, estimate the personalized price solution via a multinomial logistic model, and run a RCT to evaluate out-of-sample the effects of personalized pricing on take-up as well as overall health outcomes. 

\subsection{Research questioning the ontological stability of valuations}
 
 \paragraph{Ontological stability of valuations.} 
While a typical economic rejoinder to these concerns highlights that first-degree price discrimination achieves a Pareto-efficient allocation of goods to individuals at their valuation and universal access, we would note the actual distributional implications of personalized pricing depend on how consumer behavior does or does not reflect an implicitly assumed ontologically stable and valid idiosyncratic ``valuation''. 

While there is strong mechanism design theory for soliciting individual valuations in, for example, auctions, we argue that many of the domains where consumer-facing personalized pricing has faced pushback are different from the restricted domains where auctions are deployed in practice. We propose a taxonomy of types of failures of the ``ideal theory'' of willingness to pay which raise concerns for fairness. These settings are not merely pathological or point failures of economic theory, but rather exactly how structural inequities manifest in price optimization settings. 
    A first concern is \textit{behavioral biases} (and price optimization which extracts rent from them). For example, credit card contracts with promotional offers but high terms afterwards. This motivates the use of mandatory disclosure notices in that setting. 
    A second concern is that of \textit{differential endowments} which surface via different incomes or  credit constraints, budget constraints, and liquidity constraints. This is a concern when individuals' reflected purchase behavior is not a realization of their ideal valuation or willingness to pay, of the item, but rather of their \textit{ability} to pay. \cite{l18} discusses a similar argument against Kaldor-Hicks efficiency arguments for welfare analysis in empirical legal studies. In settings with credit constraints, individuals are credit or liquidity-constrained, and therefore borrow at high (perhaps even irrationally high) interest rates because of lack of other options. \cite{alan2013subprime}

\clearpage 
 \section{Proofs}\label{sec-proofs} 
 
 \subsection{Proofs for price parity} 
 \begin{proof}[Proof of \cref{prop-price-equity} ]
	\textbf{Optimizing over attribute-based personalized prices $p^*(x)$: }

We consider the Lagrangian dual of \cref{opt-marginal-parity}. We then applying Sion's minimax theorem to swap the order of min and max operations, which is valid under compactness \cite{boyd2004convex}: 
\begin{equation}
\min_{p(x,a)} \max_\lambda \E
\bracks{ - p D(p) + \lambda ( p \xi(A)  - \Gamma) } =\max_\lambda \min_{p(x)} \E
 \bracks{ - p D(p) + \lambda (p \xi(A)   - \Gamma) } 
  \end{equation}
For a linear demand model, observing that when $\pxa$ is unrestricted, we let $p^*(x,a; \lambda)$ be the optimal solution parameterized by $\lambda$:   
\begin{equation}\label{eqn-prop1-pxa}p^*(x,a; \lambda) \in \argmax \E[ - p D(p) + \lambda (p \xi(A)   - \Gamma)\mid X=x,A=a].\end{equation}
The optimal price with the $\lambda$ penalty is computable in closed form: $$p^*(x,a;\lambda) = \frac{- \overline{D}(x,a) + \lambda \xi(a) }{2\beta}.$$
Plugging in this solution: 
\begin{align*}
\max_\lambda L(\lambda, p^*(X,A;\lambda)) =\max_\lambda \E
\left[ -p^*(X,A;\lambda) D(p^*(X,A;\lambda)) + \lambda (p^*(X,A;\lambda) \xi - \Gamma)  \right]
\end{align*} 
We maximize the above over $\lambda$. Taking derivatives with respect to $\lambda$, we obtain the first order necessary condition for optimality, letting $p^*(\lambda) = p^*(X,A;\lambda)$, $\overline D = \overline D(X,A)$ for brevity is: 
\begin{align} 0 &=\E
\left[-p^*(\lambda)  \frac{d}{d\lambda} D(p^*(\lambda)) + -(\frac{d}{d\lambda} p^*(\lambda) )  D(p^*(\lambda)) +  (p^*(\lambda) \xi  - \Gamma)  + \lambda \frac{\xi^2}{2\beta}
\right] \nonumber\\
 &= \E
\left[-  \left(\frac{-\overline{D} + \lambda \xi }{2 \beta} \right)\frac{\xi}{2} - \frac{\xi}{2\beta} (\overline{D} + \frac 12 (-\overline{D} + \lambda \xi))
+  (p^*(\lambda) \xi  - \Gamma)  + \lambda \frac{\xi^2}{2\beta}
\right]
\nonumber \\
 &= \E
\left[ \frac{\overline{D} \xi}{4\beta} - \frac{\overline{D} \xi}{4\beta} - \lambda \frac{\xi^2}{2\beta} 
+   (p^*(\lambda) \xi  - \Gamma)  + \lambda \frac{\xi^2}{2\beta}  \right] \\
& = \E\left[ \left( \frac{- \overline{D} + \lambda \xi }{2\beta} \xi  - \Gamma \right)\right] \label{eqn-lindemand-dp-foc} 
\end{align} 
From \cref{eqn-lindemand-dp-foc} we conclude
 $$\lambda^* = \frac{\E[\overline{D}(X,A)\xi(A) + 2 \beta \Gamma]}{\E[ \xi^2(A)]} .$$ Note that further taking the derivative of eq. \ref{eqn-lindemand-dp-foc} with respect to $\lambda$ verifies $L(\lambda, p^*(\lambda))$ is concave in $\lambda$, $\frac{d^2 L(\lambda, p^*(\lambda))}{d \lambda^2 } = \E[ \frac{\xi^2}{2 \beta}]<0$ under Asn. \ref{asn-downwardsloping-lineardemand}. Therefore the first-order necessary condition is also sufficient.

 \textbf{Optimizing over attribute-blind personalized prices $p^*(x)$: }
 
 The counterpart of \cref{eqn-prop1-pxa} is, scaling by a constant $f(x)$, the covariate density of $x$, to simplify $$ \Delta f(x\mid a) \defeq \pr(X=x\mid A=a) - \pr(X=x\mid A=b) =\E[ \xi(A) \mid X=x] f(x) 
 =(\propa^{-1} \pr(A=a\mid X=x) - \propb^{-1}  \pr(A=b\mid X=x) )g(x)
 $$
\begin{align*}p^*(x; \lambda) &\in \argmin - p D(p\mid x) + \lambda (p \E[\xi(A) \mid X=x] - \Gamma)\\
p^*(x; \lambda) &\in \argmin - p D(p\mid x) f(x) + \lambda (p \E[\xi(A) \mid X=x] - \Gamma)f(x) \\
\iff p^*(x; \lambda) &\in \argmin - p D(p\mid x) f(x) + \lambda p  \Delta f(x\mid a) 
\end{align*}
Note $D(p\mid x) = D -\alpha + \E[\beta_A \mid X=x] + \E[ g(x,A) \mid x]$. 

The optimal $\lambda-$ parametrized price is 
 $$p^*(x;\lambda) = \frac{- \overline{D}(x) + \lambda \E[ \xi(A) \mid X=x]  }{2\E[ \beta_A\mid X=x]} = \frac{- \overline{D}(x) + \lambda \frac{\Delta f(x\mid a)}{f(x)} }{2\E[ \beta_A\mid X=x]}.$$
 We correspondingly solve (analogous to the previous): 
 \begin{align*}
\max_\lambda L(\lambda, p^*(X;\lambda)) =\max_\lambda \E
\left[ -p^*(X;\lambda) D(p^*(X;\lambda)) + \lambda (p^*(X;\lambda) \xi - \Gamma)  \right]
\end{align*} 
with the corresponding first-order conditions
 \begin{align} 0 &=\E
\left[-p^*(\lambda)  \frac{d}{d\lambda} D(p^*(\lambda)) + -(\frac{d}{d\lambda} p^*(\lambda) )  D(p^*(\lambda)) +  (p^*(\lambda) \xi  - \Gamma)  + \lambda \frac{\xi^2}{2 \E[ \beta_A\mid X=x] }
\right] \nonumber\\
& = \E\left[ \left( \frac{- \overline{D} + \lambda \E[ \xi(A) \mid X=x]  }{2\E[ \beta_A\mid X=x]}\E[ \xi(A) \mid X=x]   - \Gamma \right)\right] \label{eqn-lindemand-dp-foc} 
\end{align} 
Analogously we conclude
 $$\lambda^* = \frac{\E[\overline{D}(X)\E[\xi(A) \mid X] + 2 \E[ \beta_A\mid X]  \Gamma]}{\E[\E[ \xi(A) \mid X]^2]} .$$
 \qedhere
\end{proof}

    \begin{proof}[Proof of \Cref{prop-price-equity-whowins}]

    We would like to derive conditions that might inform of the sign of $p^*(x) - \pxa$. 
    
    There are a few extreme cases which might be informative (of one regime or another). 
    \begin{enumerate}
            \item $\beta_a = \beta_b,  \overline{D}(x,a) = \overline{D}(x,b)$
            
        Disparities are solely due to covariate distributions across groups. 
        \item $\beta_a > \beta_b, \overline{D}(x,a) = \overline{D}(x,b) = \overline{D}(x) $
        
        Disparities are solely due to group-level differences in price elasticity or differences in covariate distribution across groups. 
    \end{enumerate}
    We only included conclusions for case 1 in the main text. In this appendix we also provide a sufficient condition for case 2 (though this is less interpretable; hence less useful).

    Under \textbf{case 1}, 
    \begin{align}  \pxa - p^*(x) &=
    \frac{-\overline{D}(x) + \lambda^*(x,a) \xi(a)  }{\beta}
    -\frac{-\overline{D}(x) + \lambda^*(a) \E[\xi(A)\mid X=x]  }{\beta}=
    \frac{\xi(a) \lambda^*(x,a) - \E[\xi(A) \mid X=x]  \lambda^*(x)) }{\beta} \nonumber
\\
   & =  \frac{\xi(a) \lambda^*(x,a) - \xi(a) \lambda^*(x) + \xi(a) \lambda^*(x) - \E[\xi(A) \mid X=x]  \lambda^*(x)) }{\beta} \nonumber \\
    & = \frac{1}{\beta} \left( \xi(a) (\lambda^*(x,a)-\lambda^*(x)) + \lambda^*(x) (\xi(a)- \E[\xi(A) \mid X=x])  \right) \label{prop1-priceparity-situations1-eqn1}
    \end{align} 
    
    We conclude the signs of the following terms: 
    \begin{enumerate}
        \item $(\lambda^*(x,a)-\lambda^*(x)) \leq 0,$
        
            by assumption of linearity of demand, Jensen's inequality, since $f(x) = x^2$ is a convex function, and by iterated expectation: 
    \begin{equation}\label{prop1-eq-jensen} 
        \E[ \E[ \xi(A) \mid X ]^2 ] \leq \E[ \E[  \xi(A)^2 \mid X] ] = \E[ \xi(A)^2]  
    \end{equation}
    Note that $\lambda^*(x,a) \leq \lambda^*(x) \iff \E[ \E[ \xi(A) \mid X]^2 ] \leq \E[\xi(A)^2 ]  $. 
    \item $\lambda^*(x) < 0,$ under assumption for \cref{prop-price-equity} that group $a$ faces the higher unrestricted personalized prices. 
    \item $(\xi(a)- \E[\xi(A) \mid X=x]) > 0\text{ and }(\xi(b)- \E[\xi(A) \mid X=x]) < 0.$
    
    This may be verified by observing
    $$ \textstyle \frac{1}{\rho_a} > \frac{\pr(A=a\mid X=x)}{\rho_a} - \frac{\pr(A=b\mid X=x)}{\rho_b} \iff 
    1-\pr(A=a\mid X=x) > - \pr(A=b\mid X=x)\frac{\rho_a}{\rho_b}
    \iff \rho_b > - \rho_a, $$
    and concluding based on nonnegativity of $\rho_a, \rho_b$.
    \end{enumerate}

Therefore, identifying signs of terms in \cref{prop1-priceparity-situations1-eqn1}: 
\begin{align}
    \textstyle  \pxa - p^*(x) &= \textstyle
     \underbrace{\frac{1}{\beta} }_{<0}
     \left(\underbrace{ \xi(a) (\lambda^*(x,a)-\lambda^*(x))}_{\leq 0} +
     \underbrace{\lambda^*(x) }_{\leq 0}
     \underbrace{(\xi(a)- \E[\xi(A) \mid X=x])}_{\geq 0}  \right) 
     \nonumber \\
     \textstyle      p^*(x,b) - p^*(x) &= \textstyle
     \underbrace{\frac{1}{\beta} }_{<0}
     \left( \underbrace{ \xi(b)}_{<0} \underbrace{(\lambda^*(x,a)-\lambda^*(x))}_{\leq 0} +
     \underbrace{\lambda^*(x) }_{\leq 0}
     \underbrace{(\xi(b)- \E[\xi(A) \mid X=x])}_{\leq 0}  \right) \label{prop1-eqn-situation1-eqn-pbx}
\end{align}

The claim for $\pxa-p^*(x)<0$ follows by simplifying and factoring out $\rho_a$ and $\nicefrac{\lambda^*(x)}{\beta_a} > 0$: 
\begin{align*}  \pxa - p^*(x)<0
&\iff \frac{1}{\beta}   \left( \rho_a^{-1} (\lambda^*(x,a) - \lambda^*(x)) + \lambda^*(x) (\rho_a^{-1} (1-(\pr(A=a\mid X=x) - \nicefrac{\rho_a}{\rho_b}\pr(A=b\mid X=x) )\right)< 0\\
& \textstyle \iff  \left( \frac{ \E[ \E[ \xi(A)\mid X=x]^2] }{\E[ \xi(A)^2]} -1 \right) +  ( (1-(\pr(A=a\mid X=x) - \nicefrac{\rho_a}{\rho_b}\pr(A=b\mid X=x) )< 0 \\
& \textstyle \iff  \frac{ \E[ \E[ \xi(A)\mid X=x]^2] }{\E[ \xi(A)^2]}
< (\pr(A=a\mid X=x) - \nicefrac{\rho_a}{\rho_b}\pr(A=b\mid X=x) )
\end{align*} 
 The claim follows for $p^*(x,b)-p^*(x)$ directly from \cref{prop1-eqn-situation1-eqn-pbx}.

To interpret this condition, denote $\Delta f(x;a,b)$ as the \textit{covariate-driven group divergence}: 
$$ \Delta f(x;a,b) \defeq \pr(A=a\mid X=x) - \frac{\rho_a}{\rho_b}\pr(A=b\mid X=x)
= \frac{\xiax}{\rho_a^{-1}}.$$
Observe that $$  \Delta f(x;a,b) < 0 \iff \frac{\pr(A=a\mid X=x)}{\pr(A=b\mid X=x)} < \frac{\rho_a}{\rho_b} \iff \frac{\pr(X=x\mid A=a)}{\pr(X=x\mid A=b} < 1, $$
i.e. the sign of $\Delta f(x;a,b)$ depends on the covariate likelihood under the two classes (and the magnitude depends on the magnitude of this covariate-based divergence relative to the covariate-uninformed ratio, $\nicefrac{\rho_a}{\rho_b}$ ).

   Under \textbf{case 2}: 
   
       Let $\Delta \beta_a(x) = \frac{\beta_a}{\E[ \beta_A \mid X=x]}$ and recall that $\frac{\lxa}{\lx} = \frac{ \E[ \xiax^2] }{\E[ \xi(A)^2]} \leq 0$. Adding and subtracting $\frac{- \overline{D}(x) + \xi(a) \lambda^*(x)  }{\beta_a }$,
    \begin{align*} 
    & \pxa -p^*(x) > 0\\
    \iff &  \left(\frac{-\Dbarx + \lxa \xi(a)  }{\beta_a}
      -  \frac{- \Dbarx + \xi(a) \lx  }{\beta_a }\right) + 
      \left(\frac{- \Dbarx + \xi(a) \lx  }{\beta_a }
      -  \frac{- \Dbarx + \xiax\lx  }{\bexpx }\right) >0 \\
    \iff & \frac{\xi(a)}{\beta_a} (\lambda^*(x,a) - \lambda^*(x)) + -\overline{D}(x) \left( \frac{1}{\beta_a} - \frac{1}{\E[\beta_A\mid X=x]} \right) 
    +   \left( \frac{\xi(a) \lxa}{\beta_a} - \frac{\xiax \lx}{\E[\beta_A\mid X=x]} \right) > 0 \\
      \iff  &  
      {\xi(a)} \underbrace{\left( \frac{ \E[ \xiax^2] }{\E[ \xi(A)^2]}-1\right)}_{< 0 \text{ by }\cref{prop1-eq-jensen}} + -
      \frac{\overline{D}(x)}{\lx} \left( {1}- \Delta \beta_a(x) \right) 
    +  \left( \xi(a) \frac{ \E[ \xiax^2] }{\E[ \xi(A)^2]} - \xiax  \Delta \beta_a(x) \right) > 0
\end{align*}
where in the last line, we factor out $\nicefrac{\lambda^*(x)}{\beta_a} > 0$. Correspondingly for $A=a,b$ respectively: 
\begin{align*}
         \pxa -p^*(x) > 0& \iff  \underbrace{\rho_a^{-1} \left( \frac{\lxa}{\lx}-1\right)}_{< 0} + -
      \frac{\overline{D}(x)}{\lx} \left( {1}- \Delta \beta_a(x) \right) 
    + \rho_a^{-1}\left(  \frac{\lxa}{\lx} - \Delta f(x;a,b) \cdot \Delta \beta_a(x) \right) > 0
    \\
            p^*(x,b) -p^*(x) > 0& \iff  \underbrace{-\rho_b^{-1} \left( \frac{\lxa}{\lx}-1\right)}_{> 0} + -
      \frac{\overline{D}(x)}{\lx} \left( {1}-  \Delta \beta_b(x) \right) 
    +\rho_b^{-1} \left( -1\cdot \frac{\lxa}{\lx} - 
    \rho_b \xiax \Delta \beta_b(x) \right) > 0
\end{align*}

Unlike the previous case, this case does not admit determinate conclusions on signs. 

We may simplify the condition to obtain that if $ -\Delta f(x;a,b) \lx + \Dbarx \rho_a > 0,$
$$  \textstyle \pxa -p^*(x) > 0 \iff
 \Delta \beta_a(x) < \frac{ \lx -2\lxa +\Dbarx \rho_a }{ -\Delta f(x;a,b) \lx + \Dbarx \rho_a}, 
   $$
with the inequality on $\Delta \beta_a(x)$ holding in the opposite direction if instead $ -\Delta f(x;a,b) \lx + \Dbarx \rho_a <0$.
\qedhere
\end{proof}

 \subsection{Model error fairness} 
 \begin{proof}[Proof of \cref{prop-pricing-error-decomposition} ]

We omit dependence on fixed $x$ and denote $D(p) = D(p\mid x)$. Gradient $\nabla$ is with respect to $p$. We assume price elasticity of demand is nonpositive, $\nabla_p \eta< 0$. We specialize to a revenue setting by observing that $\nabla h = \eta(p) + \nabla \eta \cdot p$, so that $\hp, p^*$ satisfy the first order optimality conditions:
\begin{align*}
    \hp &= - \frac{\hat D(\hp)}{\nabla \hat D(\hp)}, \qquad p^* = - \frac{\eta(p^*)}{\nabla \eta(p^*)} .
\end{align*}

Taylor expanding $\hat D(\hp)$ around $p^*$: 
\begin{align*}
 \hat p^* - p^*    &=- \frac{\hat D(\hp)}{\nabla \hat D(\hp)} + \frac{D(p^*)}{\nabla D(p^*)}  = \frac{\heta(p^*) + \nabla \heta (p^*) (\hp - p^*) }{\nabla \hat D(\hp)} + \frac{D(p^*)}{\nabla D(p^*)} 
\end{align*}
so that 

\begin{align*}
 (\hat p^* - p^*  )\left(1 - \frac{ \nabla \heta (p^*)}{\nabla \hat D(\hp)}\right) &=  -\frac{\heta(p^*) }{\nabla \hat D(\hp)} + \frac{D(p^*)}{\nabla \hat D(\hp)} \frac{\nabla \hat D(\hp)}{\nabla D(p^*)} + o((\hp - p^*)^2)\\
 &=  -\frac{\heta(p^*) }{\nabla \hat D(\hp)} + \frac{D(p^*)}{\nabla \hat D(\hp)} 
 \left( 1 +  \frac{\nabla \hat D(\hp)- \nabla D(p^*)}{\nabla D(p^*)}  \right)+ o((\hp - p^*)^2)\\
  &=   \frac{\heta(p^*) - D(p^*)}{\nabla \hat D(\hp)}+ \frac{D(p^*)}{\nabla \hat D(\hp)} \left( \frac{\nabla \hat D(\hp)- \nabla D(p^*)}{\nabla D(p^*)}  \right)+ o((\hp - p^*)^2)
\end{align*}

Therefore, 
\begin{align*}
 (\hat p^* - p^*  )&= \left(\frac{\nabla \hat D(\hp)}{\nabla \hat D(\hp)- \nabla \heta (p^*)}\right)  
 \left(\frac{\heta(p^*) - D(p^*)}{\nabla \hat D(\hp)}+ \frac{D(p^*)}{\nabla \hat D(\hp)} \left( \frac{\nabla \hat D(\hp)- \nabla D(p^*)}{\nabla D(p^*)}  \right)
 \right) + o((\hp - p^*)^2) \\
 &= (\nabla \hat D(\hp)- \nabla \heta (p^*))^{-1}
 \left({\heta(p^*) - D(p^*)}+ \frac{D(p^*)}{\nabla D(p^*)} \left( {\nabla \hat D(\hp)- \nabla D(p^*)}  \right)
 \right) + o((\hp - p^*)^2)\\
 &
   \overset{1}{=}(\nabla \hat D(\hp)- \nabla \heta (p^*))^{-1}
 \left({\heta(p^*) - D(p^*)}+ \frac{D(p^*)}{\nabla D(p^*)} \left( {\nabla \hat D(\hp)- \nabla \heta (p^*)+ \nabla \heta (p^*)- \nabla D(p^*)}  \right)
 \right)..\\
  &=\left(\frac{{\heta(p^*) - D(p^*)}}{\nabla \hat D(\hp)- \nabla \heta (p^*)} + p^* \left( 1+ \frac{\nabla \heta (p^*)- \nabla D(p^*)}{\nabla \hat D(\hp)- \nabla \heta (p^*)} \right)
 \right) + o((\hp - p^*)^2)
\end{align*}
where in $1$ we expand $\left( {\nabla \hat D(\hp)- \nabla D(p^*)}  \right)$; then simplify.
\qedhere
\end{proof}

\subsection{Market share} 

The main tool for analyzing local sensitivities of the optimal prices is an implicit function theorem to differentiate the optimal solution with respect to the parameter. We state it for completeness. 

    \begin{theorem}[Dini Classical Implicit Function Theorem (Thm. 1B.1 of \cite{dontchev2009implicit}]
    Consider a function $f: \mathbb R^d \times \mathbb R^n \mapsto \mathbb R^n $ with values $f(\lambda, p)$ with $\lambda$ the parameter and $p$ the variable to solve for. The equation $f(\lambda,p)$ is associated with the solution mapping 
    $$ S\colon \lambda \mapsto \{ p \in \mathbb{R}^n \mid f(\lambda, p) = 0 \}, \text{ for } \lambda \in \mathbb R^d  $$
    
    Let $f$ be continuously differentiable in a neighborhood of $(\overline{\lambda}, \overline{p})$ such that $f(\overline{\lambda}, \overline{p} )=0$, and let the partial Jacobian of $f$ with respect to $p$ at $(\overline{\lambda}, \overline{p})$, namely $\nabla_x f(\overline{\lambda}, \overline{p})$. 
    
    Then the solution mapping S has a single valued localization s around $\overline{\lambda}$ for $\overline{p}$ which is continuously differentiable in a neighborhood $Q$ of $\overline{p}$ with Jacobian satisfying 
    $$ \nabla s(\lambda) = - \nabla_p f(\lambda, s(\lambda))^{-1} \nabla_\lambda f(\lambda, s(\lambda)) \text{ for every } \lambda \in Q  $$
    \end{theorem}

    Using the implicit function theorem, we can characterize the sensitivities of solutions under attribute-blind vs. attribute-based, and group market share vs. population market share penalties. We restate an expanded versio of \Cref{lemma-opt-conditions}. 
    
    \textbf{Lemma 1}[Optimality conditions for different penalties ]
    
The sensitivities of price with respect to $\lambda$, $\frac{\partial p^*(x) }{\partial \lambda}$. 

    \begin{enumerate}
    \item $p^*(x)$ with population market share penalty satisfies
        $
    \frac{1}{p^*(x)+ \lambda} + \frac{D'(p^*(x) \mid x) }{D(p^*(x)\mid x)} = 0 ,\;\; \forall x
$
so that $$\nabla_\lambda p^*(x;0) =  \frac{R''(\px\mid x))^{-1}}{\px^2}.$$
    \item $p^*(x,a)$ with population market share satisfies
        $
    \frac{1}{p^*(x,a)+ \lambda } + \frac{D'(p^*(x,a) \mid x,a) }{D(p^*(x,a)\mid x,a)} = 0 ,\;\; \forall x,a
$
so that $$\nabla_\lambda \pstar(x,a;0) =  \frac{R''(\pxa\mid x,a))^{-1}}{\pxa^2}.$$
    \item $p^*(x,a)$ with group-level market share 
    $
    \frac{1}{p^*(x,a)+\lambda_a/\rho_a} + \frac{D'(p^*(x,a) \mid x,a) }{D(p^*(x,a)\mid x,a)} = 0 
$
so that $$\nabla_\lambda p^*(x,a;0) = 
\frac{1}{\rho_a}
\frac{R''(\pxa\mid x,a))^{-1}}{\pxa^2} .$$
\end{enumerate}

    \begin{proof}[Proof of \Cref{lemma-opt-conditions}]
    \begin{enumerate}
    \item $p^*(x,a)$ with population market share
    
        \begin{align*}
        & \pxa \in \argmax \E[ p D(p)] + \lambda \E[D(p)] \\
        \iff & \pxa \in \argmax \log( (p+\lambda) D(p\mid x,a) )
    \end{align*}
    Therefore $p^*(x,a)$ satisfies the following: 
    \begin{align*}
    \frac{1}{p^*(x,a)+ \lambda } + \frac{D'(p^*(x,a) \mid x,a) }{D(p^*(x,a)\mid x,a)} = 0 ,\;\; \forall x,a
\end{align*}

The expression for $\nabla_\lambda p^*(x)(0)$ follows by applying the Implicit function theorem on the optimality condition. 
    \item $p^*(x)$ with population market share
    
        \begin{align*}
        & \px \in \argmax \E[ p D(p)] + \lambda \E[D(p)] \\
        \iff  & \px \in \argmax \E[ \E[ (p+\lambda) D(p) \mid X]  ], \forall x\\
        \iff & \px \in \argmax (p+\lambda) D(p\mid x) , \forall x\\
        \iff & \px \in \argmax \log( (p+\lambda) D(p\mid x) ), \forall x
    \end{align*}
Therefore $p^*(x)$ satisfies the following: 
    \begin{align*}
    \frac{1}{p^*(x)+\lambda} + \frac{D'(p^*(x) \mid x) }{D(p^*(x)\mid x)} = 0 ,\;\; \forall x
\end{align*}

    \item $p^*(x,a)$ with group-level market share 
        \begin{align*}
        \iff  & \pxa \in \argmax  \E[ (p+\lambda_a/\rho_a) D(p) \mid X=x,A=a]   \\
        \iff & \pxa \in \argmax (p+\lambda_a/\rho_a) D(p\mid x,a) \\
        \iff & \pxa \in \argmax \log( (p+\lambda_a/\rho_a) D(p\mid x,a) )
    \end{align*}
    
    Therefore $p^*(x,a;\lambda)$ is such that
\begin{align*}
    \frac{1}{p+\lambda_a/\rho_a} + \frac{D'(p \mid x,a) }{D(p\mid x,a)} = 0 
\end{align*}

\end{enumerate}

    \end{proof} 
    
    \subsection{Allocative efficiency: Concordance}
    
    \begin{proof}[Proof of \Cref{thm-concordance-equivalence}]
Let $\mathcal I \{ A=a\}$ denote index sets for data points within group $a$, etc. 
\begin{align*}
    \pr( D(p_i) < D(p_j) \mid P_a < P_b)  & = \frac{1}{\vert \{ (i,j) \colon p_i < p_j \}\vert } \frac{1}{n^2} \sum_{i \in \mathcal I \{ A=a\} } \sum_{j\in \mathcal I \{ A=b\}  } \mathbb{I}[ D(p_i) = 0, D(p_j) = 1, p_i < p_j  ]\\
     & = \frac{1}{\vert \{ (i,j) \colon p_i < p_j \}\vert } \frac{1}{n^2} \sum_{i \in \mathcal I \{ A=a\} } \sum_{j\in \mathcal I \{ A=b\}  } \mathbb{I}[ v_i < p_i < p_j < v_j  ]\\
     & \leq \frac{1}{\vert \{ (i,j) \colon p_i < p_j \}\vert } \frac{1}{n^2} \sum_{i \in \mathcal I \{ A=a\} } \sum_{j\in \mathcal I \{ A=b\}  } \mathbb{I}[ \{ v_i < v_j \} \cap \{ p_i < p_j  \}  ]\\
     &=  \pr(V_a > V_b \mid P_a < P_b)  
\end{align*}

where the first equality holds because under Asn.~\ref{asn-monotonicity} (a.s. monotonicity), the following events are a.s. equivalent: $$\{p_i < p_j, D(p_i) = 0, D(p_j) = 1 \} \iff v_j > v_i.$$

The second inequality holds because $\{  v_i < p_i < p_j < v_j  \} \subset 
\{ \{ v_i < v_j \} \cap \{ p_i < p_j  \}
\}$.

\end{proof}
\qedhere
 
  \section{Dataset details}\label{sec-data-details} 
  
 \paragraph{Details about the study \cite{slunge2015willingness}} We omit concerns about non-response. The survey was distributed online in 2013. $N=1116$. There are 28 data columns with information including categorical age values, gender, geographic factors and risk factors, and information about knowledge and trust about vaccines.

\end{document}